\documentclass{article}
\usepackage{arxiv}

\usepackage{soul}
\usepackage{xcolor}
\usepackage[utf8]{inputenc} % allow utf-8 input
\usepackage[T1]{fontenc}    % use 8-bit T1 fonts
\usepackage{hyperref}       % hyperlinks
\usepackage{url}            % simple URL typesetting
\usepackage{booktabs}       % professional-quality tables
\usepackage{amsfonts}       % blackboard math symbols
\usepackage{nicefrac}       % compact symbols for 1/2, etc.
\usepackage{microtype}      % microtypography
\usepackage{lipsum}		% Can be removed after putting your text content
\usepackage{graphicx}
\usepackage{doi}
\usepackage{multirow}
\usepackage{multicol}
\usepackage{xcolor}
\usepackage{subcaption}
\usepackage[normalem]{ulem}
\usepackage{amsmath,amsfonts,bm}
\usepackage{amssymb,amsthm}
\usepackage{enumitem}
\usepackage{subcaption}
\usepackage{adjustbox}
\usepackage{afterpage}
\usepackage[
    backend=biber,
    citestyle=authoryear,
    bibstyle=authortitle,
    mincitenames=1,
    maxcitenames=1,
    maxbibnames=3
]{biblatex}
\newcommand{\ourname}{{Kimi-VL}}
\newcommand{\ourreasoningname}{{\ourname-Thinking}}

\newcommand{\gpto}[1]{\textcolor{gray}{#1}}

\newcommand{\citep}[1]{\parencite{#1}}

\setlist[itemize,1]{leftmargin=\dimexpr 18pt}
\setlist[enumerate,1]{leftmargin=\dimexpr 18pt}

\title{
\raisebox{-0.1\height}{\includegraphics[width=0.032\textwidth]{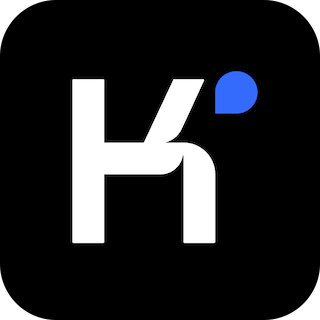}} %
\textsc{\ourname~}Technical Report}

\author{Kimi Team}

\date{}

\renewcommand{\headeright}{\ourname~Technical Report}
\renewcommand{\undertitle}{}
\renewcommand{\shorttitle}{\raisebox{-0.12\height}{\includegraphics[width=0.02\textwidth]{figures/logo.png}}}

\begin{document}
\maketitle

\vspace{-10pt}
\begin{abstract}

We present \textbf{\ourname}, an efficient open-source Mixture-of-Experts (MoE) vision-language model (VLM) that offers \textbf{advanced multimodal reasoning, long-context understanding, and strong agent capabilities}—all while activating only \textbf{2.8B} parameters in its language decoder (Kimi-VL-A3B).

\ourname~demonstrates strong performance across challenging domains:
as a general-purpose VLM, \ourname~excels in multi-turn agent tasks (\textit{e.g.,}~OSWorld), matching flagship models.
Furthermore, it exhibits remarkable capabilities across diverse challenging vision language tasks, including college-level image and video comprehension, OCR, mathematical reasoning, multi-image understanding. 
In comparative evaluations, it effectively competes with cutting-edge efficient VLMs such as GPT-4o-mini, Qwen2.5-VL-7B, and Gemma-3-12B-IT, while surpassing GPT-4o in several key domains.

\ourname{} also advances in processing long contexts and perceiving clearly. With a 128K extended context window, \ourname~can process diverse long inputs, achieving impressive scores of 64.5 on LongVideoBench and 35.1 on MMLongBench-Doc. Its native-resolution vision encoder, MoonViT, further allows it to see and understand ultra-high-resolution visual inputs, achieving 83.2 on InfoVQA and 34.5 on ScreenSpot-Pro, while maintaining lower computational cost for common tasks.

Building upon \ourname, we introduce an advanced long-thinking variant: \textbf{\ourname-Thinking-2506}. Developed through long chain-of-thought (CoT) supervised fine-tuning (SFT) and reinforcement learning (RL), the latest model exhibits strong long-horizon reasoning capabilities (64.0 on MMMU, 46.3 on MMMU-Pro, 56.9 on MathVision, 80.1 on MathVista, 65.2 on VideoMMMU) while obtaining robust general abilities (84.4 on MMBench, 83.2 on V* and 52.8 on ScreenSpot-Pro). With only around 3B activated parameters, it sets a new standard for efficient yet capable multimodal \textit{thinking} models.
Code and models are publicly accessible at \url{https://github.com/MoonshotAI/Kimi-VL}.

\end{abstract}

\begin{figure}[htb!]
    \centering
    \includegraphics[width=0.75\textwidth]{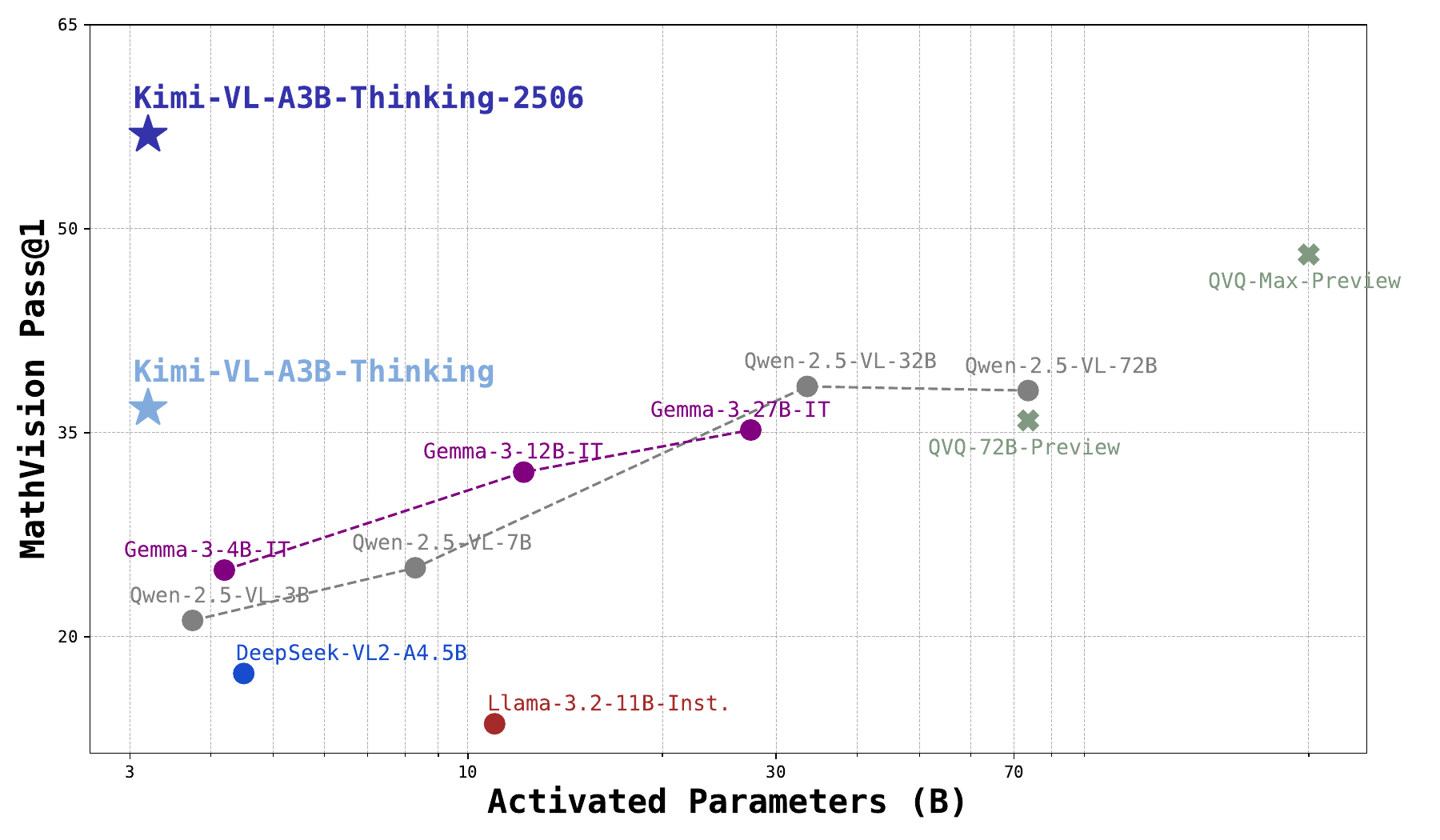}
    
    \caption{Comparison between \textbf{\ourreasoningname-2506}~and frontier open-source VLMs, including short-thinking VLMs (\textit{e.g.}~Gemma-3 series, Qwen2.5-VL series) and long-thinking VLMs (QVQ-72B/Max-Preview), on MathVision benchmark. Our model achieves strong multimodal reasoning with just 2.8B LLM activated parameters.}
    \label{fig:results_via_parameter}
\end{figure}

\begin{figure}[htb]
    \centering
    %\vspace{-24pt}
    \adjustbox{center}{%
        \includegraphics[width=0.96\textwidth]{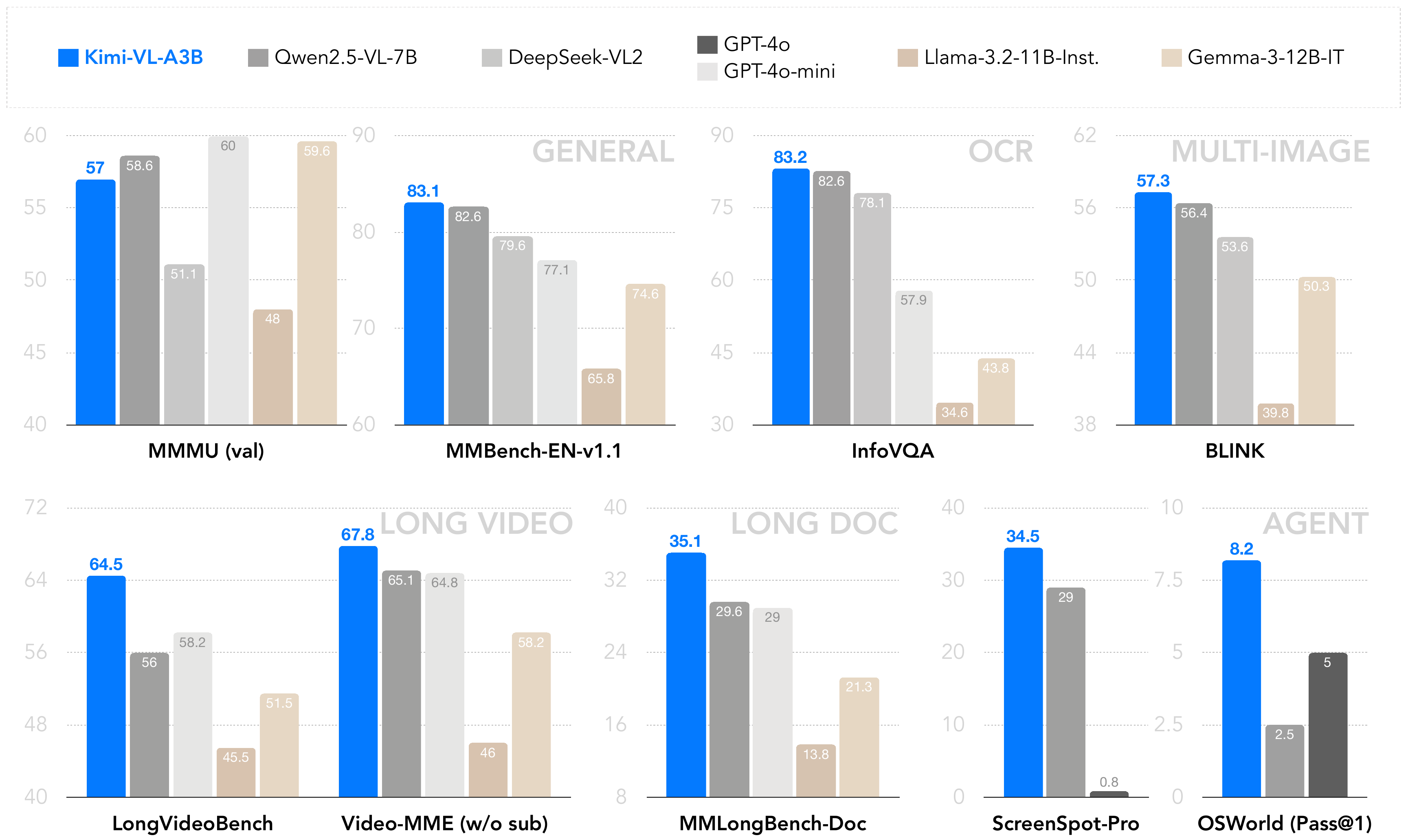}
    }
    \caption{Highlights of \textbf{\ourname}~performance for a wide range of benchmarks like, general benchmarks (MMMU, MMBench), OCR (InfoVQA), multi-image (BLINK), long video (LongVideoBench, Video-MME), long document (MMLongBench-Doc), and agent (ScreenSpot-Pro and OSWorld). Detailed results are presented in Table~\ref{tab:short_perf}.}
    \label{fig:General-results}
\end{figure}

\section{Introduction}

With the rapid advancement of artificial intelligence, 
human expectations for AI assistants have transcended traditional language-only interactions, 
increasingly aligning with the inherently multimodal nature of our world. 
To better understand and interact with these expectations, new generations of natively multimodal models, such as GPT-4o~\citep{openai2024gpt4ocard} and Google Gemini~\citep{geminiteam2024gemini15unlockingmultimodal}, have emerged with the capability to seamlessly perceive and interpret visual inputs alongside language processing.
Most recently, advanced multimodal models, pioneered by OpenAI o1 series~\citep{o12024} and Kimi k1.5~\citep{team2025kimi}, have further pushed these boundaries by incorporating deeper and longer reasoning on multimodal inputs, thereby tackling more complex problems in the multimodal domain.

Nevertheless, development in large VLMs in the open-source community has significantly lagged behind their language-only counterparts, particularly in aspects of scalability, computational efficiency, and advanced reasoning capabilities. While language-only model DeepSeek R1~\citep{deepseekai2025deepseekr1incentivizingreasoningcapability} has already leveraged the efficient and more scalable mixture-of-experts (MoE) architecture and facilitated sophisticated long chain-of-thought (CoT) reasoning, most recent open-source VLMs, \textit{e.g.}~Qwen2.5-VL~\citep{bai2025qwen25vltechnicalreport} and Gemma-3~\citep{gemmateam2025gemma3technicalreport}, continue to rely on dense architectures and do not support long-CoT reasoning. Early explorations into MoE-based vision-language models, such as DeepSeek-VL2~\citep{wu2024deepseekvl2mixtureofexpertsvisionlanguagemodels} and Aria~\citep{li2024ariaopenmultimodalnative}, exhibit limitations in other crucial dimensions. Architecturally, both models still adopt relatively traditional fixed-size vision encoders, hindering their adaptability to diverse visual inputs. From a capability perspective, DeepSeek-VL2 supports only a limited context length (4K), while Aria falls short in fine-grained visual tasks. Additionally, neither of them supports long-thinking abilities. Consequently, there remains a pressing need for an open-source VLM that effectively integrates structural innovation, stable capabilities, and enhanced reasoning through long-thinking.

In light of this, we present \textbf{\ourname}, a vision-language model for the open-source community. Structurally, \ourname{} consists of our Moonlight~\citep{liu2025muonscalablellmtraining} MoE language model with only \textbf{2.8B} activated (16B total) parameters, paired with a 400M native-resolution MoonViT vision encoder. In terms of capability, as illustrated in Figure~\ref{fig:General-results}, \ourname{} can robustly handle diverse tasks (fine-grained perception, math, college-level problems, OCR, agent, \textit{etc.}) across a broad spectrum of input forms (single-image, multi-image, video, long-document, \textit{etc.}). Specifically, it features the following exciting abilities:

\textbf{1)} \textbf{\ourname~is smart}: it has comparable text ability against efficient pure-text LLMs; without long thinking, \ourname~is already competitive in multimodal reasoning and multi-turn agent benchmarks, \textit{e.g.,~}MMMU, MathVista, OSWorld.

\textbf{2)} \textbf{\ourname~processes long}: it effectively tackles long-context understanding on various multimodal inputs within its 128K context window, far ahead of similar-scale competitors on long video benchmarks and MMLongBench-Doc.

\textbf{3)} \textbf{\ourname~perceives clear}: it shows all-round competitive ability over existing efficient dense and MoE VLMs in various vision-language scenarios: visual perception, visual world knowledge, OCR, high-resolution OS screenshot, \textit{etc.}

Furthermore, with long-CoT activation and reinforcement learning (RL), we introduce the long-thinking version of \ourname, \textbf{\ourreasoningname}, which further substantially improves performance on more complex multimodal reasoning scenarios. Despite its small scale, \ourreasoningname{} offers compelling performance on hard reasoning benchmarks (\textit{e.g.,~}MMMU, MathVision, MathVista), outperforming many state-of-the-art VLMs with even larger sizes. We further release and improved version of the thinking model, \textbf{\ourreasoningname-2506}. The improved version has even better performance on these reasoning benchmarks while retaining or improving on common visual perception and understanding scenarios, \textit{e.g.} high-resolution perception (V*), OS grounding, video and long document understanding.

\section{Approach}

\subsection{Model Architecture}

\begin{figure}
    \centering
    \includegraphics[width=0.9\linewidth]{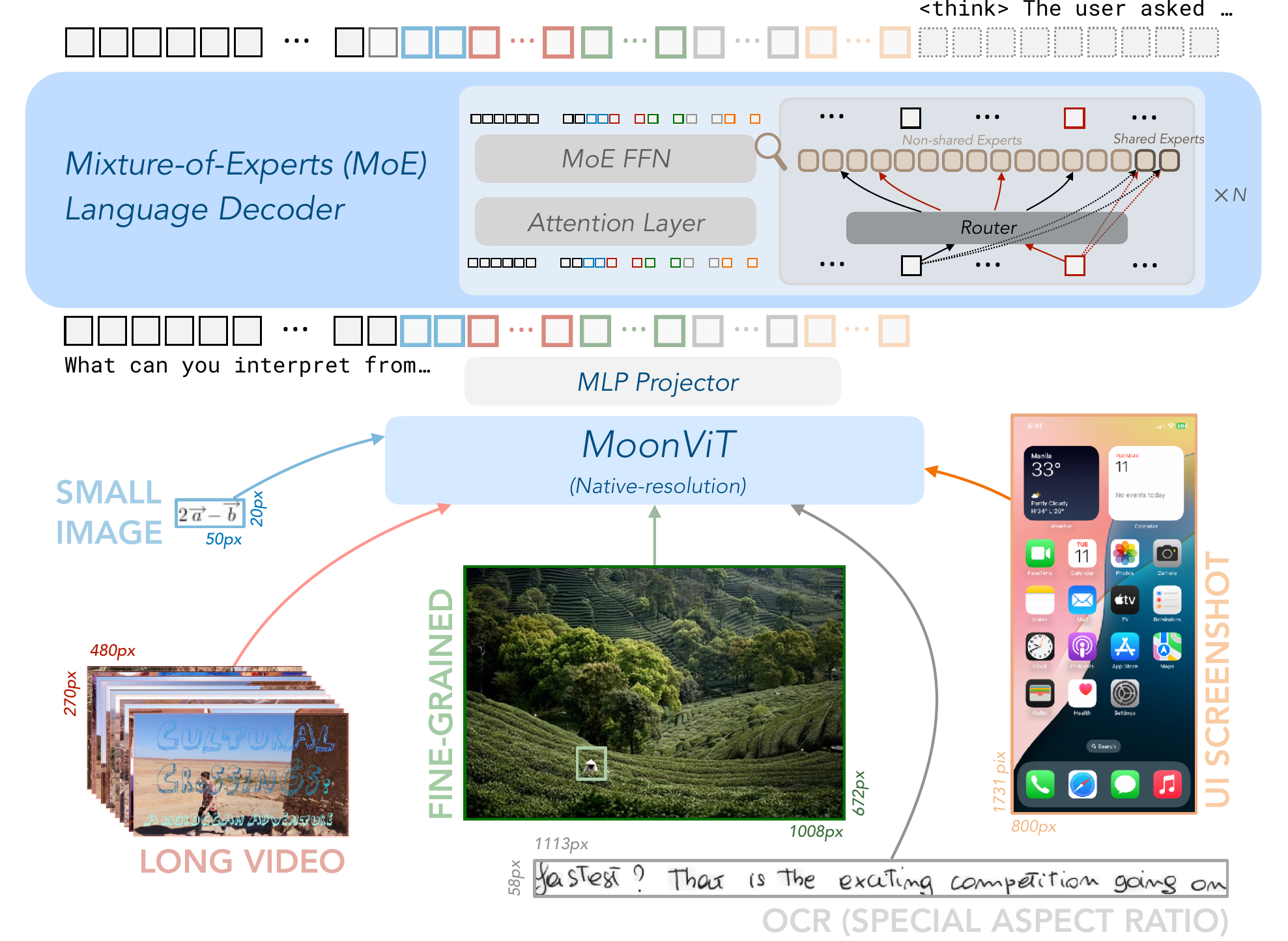}
    \vspace{-0.5em}
    \caption{The model architecture of \ourname~and \ourreasoningname, consisting of a MoonViT that allows native-resolution images, an MLP projector, and a Mixture-of-Experts (MoE) language decoder.}
    \label{fig:arch}
\end{figure}

The architecture of \ourname~consists of three parts: a native-resolution vision encoder (MoonViT), an MLP projector, and an MoE language
model, as depicted in Figure~\ref{fig:arch}. We introduce each part in this section.

\paragraph{MoonViT: A Native-resolution Vision Encoder} %\ychen{blue}{increase the length here}
We design MoonViT, the vision encoder of \ourname, to natively process images at their varying resolutions, eliminating the need for complex sub-image splitting and splicing operations, as employed in LLaVA-OneVision~\citep{li2024llavaonevisioneasyvisualtask}. We incorporate the packing method from NaViT~\citep{dehghani2023patchnpacknavit}, where images are divided into patches, flattened, and sequentially concatenated into 1D sequences. These preprocessing operations enable MoonViT to share the same core computation operators and optimization as a language model, such as the variable-length sequence attention mechanism supported by FlashAttention~\citep{dao2022flashattentionfastmemoryefficientexact}, ensuring non-compromised training throughput for images of varying resolutions.

MoonViT is initialized from and continually pre-trained on SigLIP-SO-400M~\citep{zhai2023sigmoidlosslanguageimage}, which originally employs learnable fixed-size absolute positional embeddings to encode spatial information. While we interpolate these original position embeddings to better preserve SigLIP's capabilities, these interpolated embeddings become increasingly inadequate as image resolution increases. To address this limitation, we incorporate 2D rotary positional embedding (RoPE)~\citep{su2023roformerenhancedtransformerrotary} across the height and width dimensions, which improves the representation of fine-grained positional information, especially in high-resolution images. These two positional embedding approaches work together to encode spatial information for our model and seamlessly integrate with the flattening and packing procedures. This integration enables MoonViT to efficiently process images of varying resolutions within the same batch. The resulting continuous image features are then forwarded to the MLP projector and, ultimately, to the MoE language model for subsequent training stages. In Kimi-VL-A3B-Thinking-2506, we further continually train this MoonViT to authentically encode up to 3.2 million pixels from a single image, 4 times compared to the original limit.

\paragraph{MLP Projector}
We employ a two-layer MLP to bridge the vision encoder (MoonViT) and the LLM. Specifically, we first use a pixel shuffle operation to compress the spatial dimension of the image features extracted by MoonViT, performing 2×2 downsampling in the spatial domain and correspondingly expanding the channel dimension. We then feed the pixel-shuffled features into a two-layer MLP to project them into the dimension of LLM embeddings.

\paragraph{Mixture-of-Experts (MoE) Language Model} The language model of \ourname~utilizes our Moonlight model~\citep{liu2025muonscalablellmtraining}, an MoE language model with 2.8B activated parameters, 16B total parameters, and an architecture similar to DeepSeek-V3~\citep{deepseekai2025deepseekv3technicalreport}. For our implementation, we initialize from an intermediate checkpoint in Moonlight's pre-training stage—one that has processed 5.2T tokens of pure text data and activated an 8192-token (8K) context length. We then continue pre-training it using a joint recipe of multimodal and text-only data totaling 2.3T tokens, as detailed in Sec.~\ref{sec:pretraining_stages}.

\subsection{Muon Optimizer}

We use an enhanced Muon optimizer \citep{liu2025muon}  for model optimization. Compared to the original Muon optimizer \citep{jordan2024muon}, we add weight decay and carefully adjust the per-parameter update scale. Additionally, we develop a distributed implementation of Muon following the ZeRO-1 \citep{rajbhandari2020zero} optimization strategy, which achieves optimal memory efficiency and reduced communication overhead while preserving the algorithm’s mathematical properties. This enhanced Muon optimizer is used throughout the entire training process to optimize all model parameters, including the vision encoder, the projector, and the language model.

\subsection{Pre-Training Stages}
\label{sec:pretraining_stages}

\begin{figure}[t]
\centering
\includegraphics[width=\textwidth]{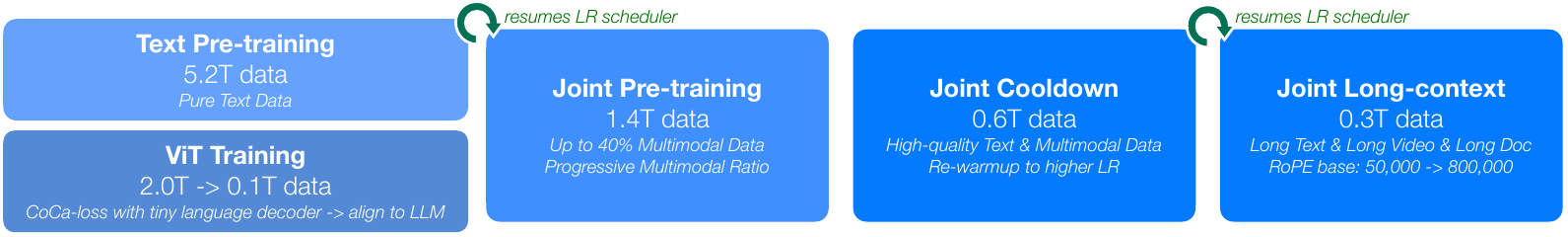}
\caption{The pre-training stages of \ourname{} consume a total of 4.4T tokens after text-only pre-training of its language model. To preserve text abilities, all stages that update the language model are joint training stages.}
\label{fig:pretrainingstages}
\end{figure}

As illustrated in Figure~\ref{fig:pretrainingstages} and Table~\ref{tab:pretrainingdatavolume}, after loading the intermediate language model discussed above, \ourname's pre-training comprises a total of 4 stages consuming 4.4T tokens overall: first, standalone ViT training to establish a robust native-resolution visual encoder, followed by three joint training stages (pre-training, cooldown, and long-context activation) that simultaneously enhance the model's language and multimodal capabilities. The details are as follows.

\begin{table}[h]
\centering
\caption{Overview of training stages: data composition, token volumes, sequence lengths, and trainable components.}
\vspace{0.3em}
\begin{tabular}{lcccc}
\toprule
\textbf{Stages} & \textbf{ViT Training} & \textbf{Joint Pre-training} & \textbf{Joint Cooldown} & \textbf{Joint Long-context} \\
\midrule
\textbf{Data} & Alt text & + & + & + \\
 & Synthesis Caption & Text, Knowledge & High-quality Text & Long Text \\
 & Grounding & Interleaving & High-quality Multimodal & Long Video \\
 & OCR & Video, Agent & Academic Sources & Long Document \\
\midrule
\textbf{Tokens} & 2T + 0.1T & 1.4T & 0.6T & 0.3T \\
\midrule
\textbf{Sequence length} & 8192 & 8192 & 8192 & 32768->131072 \\
\midrule
\textbf{Training} & ViT & ViT \& LLM & ViT \& LLM& ViT \& LLM \\
\bottomrule
\end{tabular}
\label{tab:pretrainingdatavolume}
\end{table}

\paragraph{ViT Training Stages}

The MoonViT is trained on image-text pairs, where the text components consist of a variety of targets: image alt texts, synthetic captions, grounding bboxes, and OCR texts. The training incorporates two objectives: a SigLIP~\citep{zhai2023sigmoidlosslanguageimage} loss $\mathcal{L}_{siglip}$ (a variant of contrastive loss) and a cross-entropy loss $\mathcal{L}_{caption}$ for caption generation conditioned on input images. Following CoCa's approach~\citep{yu2022cocacontrastivecaptionersimagetext}, the final loss function is formulated as $\mathcal{L}=\mathcal{L}_{siglip}+\lambda\mathcal{L}_{caption}$, where $\lambda=2$. Specifically, the image and text encoders compute the contrastive loss, while the text decoder performs next-token prediction (NTP) conditioned on features from the image encoder. To accelerate training, we initialized both encoders with SigLIP SO-400M~\citep{zhai2023sigmoidlosslanguageimage} weights and implemented a progressive resolution sampling strategy to gradually allow larger size; the text decoder is initialized from a tiny decoder-only language model. During training, we observed an emergence in the caption loss while scaling up OCR data, indicating that the text decoder had developed some OCR capabilities. After training the ViT in the CoCa-alike stage with 2T tokens, we align the MoonViT to the MoE language model using another 0.1T tokens, where only MoonViT and MLP projector are updated. This alignment stage significantly reduces the initial perplexity of MoonViT embeddings in the language model, allowing a smoother joint pre-training stage as follows.

\paragraph{Joint Pre-training Stage} In the joint pre-training stage, we train the model with a combination of pure text data (sampled from the same distribution as the initial language model) and a variety of multimodal data (as discussed in Sec.~\ref{sec:pretraindata}). We continue training from the loaded LLM checkpoint using the same learning rate scheduler, consuming an additional 1.4T tokens. The initial steps utilize solely language data, after which the proportion of multimodal data gradually increases. Through this progressive approach and the previous alignment stage, we observe that joint pre-training preserves the model's language capabilities while successfully integrating visual comprehension abilities.

\paragraph{Joint Cooldown Stage}

The stage following the pre-training stage is a multimodal cooldown phase, where the model is continue trained with high-quality language and multimodal datasets to ensure superior performance. 
For the language part, through empirical investigation, we observe that the incorporation of synthetic data during the cooling phase yields significant performance improvements, particularly in mathematical reasoning, knowledge-based tasks, and code generation. The general text components of the cooldown dataset are curated from high-fidelity subsets of the pre-training corpus. For math, knowledge, and code domains, we employ a hybrid approach: utilizing selected pre-training subsets while augmenting them with synthetically generated content. Specifically, we leverage existing mathematical knowledge and code corpora as source material to generate question-answer (QA) pairs through a proprietary language model, implementing rejection sampling techniques to maintain quality standards~\citep{yue2023mammoth,su2024nemotron}. These synthesized QA pairs undergo comprehensive validation before being integrated into the cooldown dataset. For the multimodal part, in addition to the two strategies as employed in text cooldown data preparation, \textit{i.e.}~question-answer synthesis and high-quality subset replay, to allow more comprehensive visual-centric perception and understanding~\citep{li2024llavaonevisioneasyvisualtask,tong2024cambrian1fullyopenvisioncentric,guo2024mammothvlelicitingmultimodalreasoning}, we filter and rewrite a variety of academic visual or vision-language data sources to QA pairs. Unlike post-training stages, these language and multimodal QA pairs in the cooldown stage are only included for activating specific abilities and henceforth facilitating learning high-quality data, thus, we keep their ratio at a low portion to avoid overfitting these QA patterns. The joint cooldown stage significantly improves both language and multimodal abilities of the model.

\begin{table}[t!]
\centering
\caption{\textbf{Needle-in-a-Haystack (NIAH)} test on text/video haystacks, where needles are uniformly distributed at various positions within the haystack. We report recall accuracy across different haystack lengths up to 131,072 tokens (128K).}
\vspace{0.3em}
\resizebox{\linewidth}{!}{\begin{tabular}{l|ccccccc}
\toprule
Haystack Length             & \small{$(0, 2048]$} & \small{$(2048, 4096]$} & \small{$(4096, 8192]$} & \small{$(8192, 16384]$} & \small{$(16384, 32768]$} & \small{$(32768, 65536]$} & \small{$(65536, 131072]$} \\
\midrule
\textit{- text haystack}  &   100.0    &  100.0  &  100.0  &   100.0  &  100.0   &  100.0   &  87.0    \\
\textit{- video haystack} &   100.0    &  100.0  &  100.0  &   100.0  &  100.0   &  100.0   &   91.7  \\
\bottomrule
\end{tabular}}
\label{niahtest}
\end{table}

\paragraph{Joint Long-context Activation Stage}

In the final pre-training stage, we extend the context length of the model from 8192 (8K) to 131072 (128K), with the inverse frequency of its RoPE~\citep{su2023roformerenhancedtransformerrotary} embeddings reset from 50,000 to 800,000. The joint long-context stage is conducted in two sub-stages, where each one extends the model's context length by four times. For data composition, we filter and upsample the ratio of long data to 25\% in each sub-stage, while using the remaining 75\% tokens to replay shorter data in its previous stage; our exploration confirms that this composition allows the model to effectively learn long-context understanding while maintaining short-context ability. 

To allow the model to activate long-context abilities on both pure-text and multimodal inputs, the long data used in \ourname's long-context activation consists of not only long text, but also long multimodal data, including long interleaved data, long videos, and long documents. Similar as cooldown data, we also synthesize a small portion of QA pairs to augment the learning efficiency of long-context activation. After the long-context activations, the model can pass needle-in-a-haystack (NIAH) evaluations with either long pure-text or long video haystack, proving its versatile long-context ability. We provide the NIAH recall accuracy on various range of context length up to 128K in Table~\ref{niahtest}.

\begin{figure}[t]
\centering
\includegraphics[width=\textwidth]{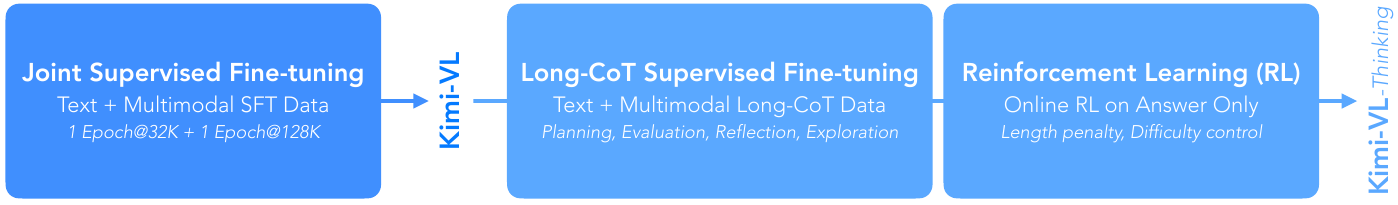}
\caption{The post-training stages of \ourname~and \ourreasoningname, including two stages of joint SFT in 32K and 128K context, and further long-CoT SFT and RL stages to activate and enhance long thinking abilities.}
\label{fig:posttrainingstages}
\end{figure}

\subsection{Post-Training Stages}

\paragraph{Joint Supervised Fine-tuning (SFT)}

In this phase, we fine-tune the base model of \ourname~with instruction-based fine-tuning to enhance its ability to follow instructions and engage in dialogue, culminating in the creation of the interactive \ourname~model. This is achieved by employing the ChatML format (Openai, 2024), which allows for a targeted instruction optimization while maintaining architectural consistency with \ourname. We optimize the language model, MLP projector, and vision encoder using a mixture of pure-text and vision-language SFT data, which will be described in Sec~\ref{sec:3.2}. Supervision is applied only to answers and special tokens, with system and user prompts being masked. The model is exposed to a curated set of multimodal instruction-response pairs, where explicit dialogue role tagging, structured injection of visual embeddings, and preservation of cross-modal positional relationships are ensured through the format-aware packing. Additionally, to guarantee the model's comprehensive proficiency in dialogue, we incorporate a mix of multimodal data and pure text dialogue data used in Moonlight, ensuring its versatility across various dialogue scenarios.

We first train the model at the sequence length of 32k tokens for 1 epoch, followed by another epoch at the sequence length of 128k tokens. In the first stage (32K), the learning rate decays from $2\times10^{-5}$ to $2\times10^{-6}$, before it re-warmups to $1\times10^{-5}$ in the second stage~(128K) and finally decays to $1\times10^{-6}$. To improve training efficiency, we pack multiple training examples into each single training sequence.

\paragraph{Long-CoT Supervised Fine-Tuning}

\label{sec:long_cot_sft}

With the refined RL prompt set, we employ prompt engineering to construct a small yet high-quality long-CoT warmup dataset, containing accurately verified reasoning paths for both text and image inputs. 
This approach resembles rejection sampling (RS) but focuses on generating long-CoT reasoning paths through prompt engineering. 
The resulting warmup dataset is designed to encapsulate key cognitive processes that are fundamental to human-like reasoning, such as \textbf{planning}, where the model systematically outlines steps before execution; \textbf{evaluation}, involving critical assessment of intermediate steps; \textbf{reflection}, enabling the model to reconsider and refine its approach; and \textbf{exploration}, encouraging consideration of alternative solutions.
By performing a lightweight SFT on this warm-up dataset, we effectively prime the model to internalize these multimodal reasoning strategies. As a result, the fine-tuned long-CoT model demonstrates improved capability in generating more detailed and logically coherent responses, which enhances its performance across diverse reasoning tasks.

\paragraph{Reinforcement Learning}

To further advance the model’s reasoning abilities, we then train the model with reinforcement learning (RL), enabling the model to autonomously generate structured CoT rationales.
Specifically, similar as Kimi k1.5~\citep{team2025kimi}, we adopt a variant of online policy mirror descent as our RL algorithm, which iteratively refines the policy model $\pi_\theta$ to improve its problem-solving accuracy.
During the $i$-th training iteration, we treat the current model as a reference policy model and optimize the following objective, regularized by relative entropy to stabilize policy updates:
\begin{align}
\max_\theta \mathbb{E}_{(x, y^*)\sim\mathcal{D}}\left[ \mathbb{E}_{(y, z)\sim\pi_\theta} \left[r(x, y, y^*)\right] - \tau \mathrm{KL} (\pi_{\theta}(x) || \pi_{\theta_i}(x))  \right]\, ,
\label{eq:opmd}
\end{align}
where $r$ is a reward model that justifies the correctness of the proposed answer $y$ for the given problem $x$, by assigning a value $r(x, y, y^*)\in \{0,1\}$ based on the ground truth $y^*$, and  $\tau > 0$ is a parameter controlling the degree of regularization.

Each training iteration begins by sampling a problem batch from the dataset $\mathcal{D}$, and the model parameters are updated to $\theta_{i+1}$ using the policy gradient derived from \eqref{eq:opmd}, with the optimized policy model subsequently assuming the role of reference policy for the subsequent iteration.
To enhance RL training efficiency, we implement a length-based reward to penalize excessively long responses, mitigating the overthinking problem where the model generates redundant reasoning chains.
Besides, we employ two sampling strategies including curriculum sampling and prioritized sampling, which leverage difficulty labels and per-instance success rates to focus training effort on the most pedagogically valuable examples, thereby optimizing the learning trajectory and improving training efficiency.

Through large-scale reinforcement learning training, we can derive a model that harnesses the strengths of both basic prompt-based CoT reasoning and sophisticated planning-enhanced CoT approaches. During inference, the model maintains standard autoregressive sequence generation, eliminating the deployment complexities associated with specialized planning algorithms that require parallel computation.
Simultaneously, the model develops essential meta-reasoning abilities including error detection, backtracking, and iterative solution refinement by effectively utilizing the complete history of explored reasoning paths as contextual information.
With endogenous learning from its complete reasoning trace history, the model can effectively encode planned search procedures into its parametric knowledge.

\subsection{Infrastructure}

\textbf{Storage}
We utilize S3 \citep{amazon_s3} compatible object storage from cloud service vendors to store our visual-text data.
To minimize the time between data preparation and model training, we store visual data in its original format and have developed an efficient and flexible data loading system.
This system provides several key benefits:
\begin{itemize}%[noitemsep,topsep=0pt]
\item Supports on-the-fly data shuffling, mixing, tokenization, loss masking and packing during training, allowing us to adjust data proportions as needed;
\item Enables random augmentation of both visual and text data, while preserving the correctness of 2D coordinate and orientation information during transformations;
\item Ensures reproducibility by strictly controlling random states and other states across different data loader workers, guaranteeing that any interrupted training can be resumed seamlessly—the data sequence after resumption remains identical to an uninterrupted run;
\item Delivers high-performance data loading: through multiple caching strategies, our system reliably supports training on large scale clusters while maintaining controlled request rates and throughput to the object storage.
\end{itemize}
Additionally, to ensure consistent dataset quality control, we developed a centralized platform for data registration, visualization, compiling statistics, synchronizing data across cloud storage systems, and managing dataset lifecycles.

\begin{figure}
    \centering
    \includegraphics[width=1.0\linewidth]{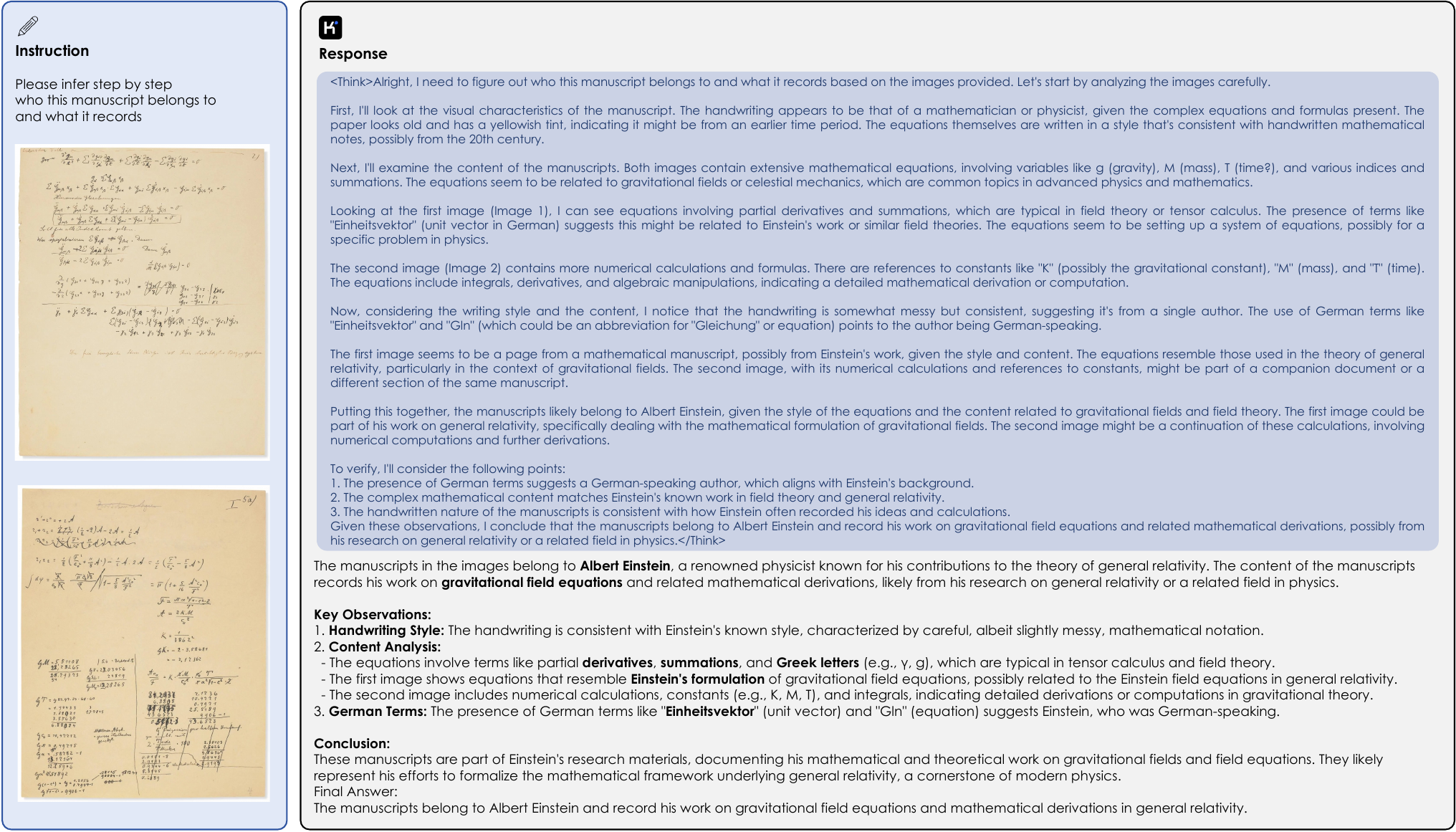}
    \caption{Manuscript reasoning visualization. \ourreasoningname~demonstrates the ability to perform historical and scientific inference by analyzing handwritten manuscripts step by step. In this example, our model identifies the author as Albert Einstein based on handwriting style, content analysis, and language cues. It reasons that the manuscripts relate to gravitational field equations, consistent with Einstein’s contributions to general relativity.}
    \label{fig:outline}
\end{figure}

\textbf{Parallelism} We adopt a 4D parallelism strategy—{Data Parallelism}~\citep{li2020pytorchdistributedexperiencesaccelerating}, 
{Expert Parallelism}~\citep{fedus2022switchtransformersscalingtrillion}, 
{Pipeline Parallelism}~\citep{huang2019gpipeefficienttraininggiant,narayanan2021efficientlargescalelanguagemodel}, 
and {Context Parallelism}~\citep{jacobs2023deepspeedulyssesoptimizationsenabling,liu2023ringattentionblockwisetransformers}—to accelerate the speed of {\ourname~}. After optimizing parallel strategies, the resulting training throughput of our model is around 60\% higher than a 7B dense VLM (\textit{e.g.}~VLMs based on Qwen2.5-7B).

\begin{itemize}
    \item \textbf{Data Parallelism (DP).} 
    DP replicates the model across multiple devices, each processing different micro-batches. 
    This setup allows larger effective batch sizes by simply increasing the number of devices.

    \item \textbf{Expert Parallelism (EP).}
    EP distributes expert modules in the MoE layer across multiple devices. 
    When combined with DP, experts on a given device can handle tokens from different DP groups, enhancing computational efficiency.

    \item \textbf{Pipeline Parallelism (PP).}
    PP splits the model into multiple layer-based stages. To minimize pipeline bubbles, we allocate the Vision Tower (VT) and several decoder layers to the first stage, place the output layer and additional decoder layers in the last stage, and distribute the remaining decoder layers evenly across intermediate stages based on their time overhead.

    \item \textbf{Context Parallelism (CP).}
    CP addresses long-sequence training by splitting sequences across different CP ranks in conjunction with flash attention~\citep{dao2022flashattentionfastmemoryefficientexact}. 
    This substantially reduces peak memory usage and relieves the memory pressure from attention computations.
\end{itemize}

Beyond these four parallel strategies, we incorporate ZeRO1~\citep{rajbhandari2020zero} and Selective Checkpointing Activation~\citep{chen2016trainingdeepnetssublinear, korthikanti2022reducingactivationrecomputationlarge} to further optimize memory usage. ZeRO1 reduces optimizer state overhead by using a distributed optimizer while avoiding extra communication costs. Selective Checkpointing Activation trades time for space by recomputing only those layers that have low time overhead but high memory consumption, striking a balance between computation efficiency and memory demands. For extremely long sequences, we expand recomputation to a broader set of layers to prevent out-of-memory errors.

\section{Data Construction}

\label{sec:data}

\subsection{Pre-Training Data}
\label{sec:pretraindata} % s

Our multimodal pre-training corpus is designed to provide high-quality data that enables models to process and understand information from multiple modalities, including text, images, and videos. 
To this end, we have also curated high-quality data from six categories -- caption, interleaving, OCR, knowledge, video, and agent -- to form the corpus.

When constructing our training corpus, we developed several multimodal data processing pipelines to ensure data quality, encompassing filtering, synthesis, and deduplication. 
Establishing an effective multimodal data strategy is crucial during the joint training of vision and language, as it both preserves the capabilities of the language model and facilitates alignment of knowledge across diverse modalities.

We provide a detailed description of these sources in this section, which is organized into the following categories:

\paragraph{Caption Data}
Our caption data provides the model with fundamental modality alignment and a broad range of world knowledge. By incorporating caption data, the multimodal LLM gains wider world knowledge with high learning efficiency. We have integrated various open-source Chinese and English caption datasets like \citep{schuhmann2022laion, gadre2024datacomp} and also collected substantial in-house caption data from multiple sources. However, throughout the training process, we strictly limit the proportion of synthetic caption data to mitigate the risk of hallucination stemming from insufficient real-world knowledge.

For general caption data, we follow a rigorous quality control pipeline that avoids duplication and maintain high image-text correlation. We also vary image resolution during pre-training to ensure that the vision tower remains effective when processing images of both high- and low-resolution.

\textbf{Image-text Interleaving Data}
During the pre-training phase, the model benefits from interleaving data for many aspects. For example, multi-image comprehension ability can be boosted by interleaving data; interleaving data always provides detailed knowledge for the given image; a longer multimodal context learning ability can also be gained by interleaving data. What's more, we also find that interleaving data can contribute positively to maintaining the model’s language abilities. 
Thus, image-text interleaving data is an important part in our training corpus. 
Our multimodal corpus considered open-sourced interleave datasets like ~\citep{zhu2024multimodal,laurenccon2024obelics} and also constructed large-scale in-house data using resources like textbooks, webpages, and tutorials.
Further, we also find that synthesizing the interleaving data benefits the performance of multimodal LLM for keeping the text knowledge.  
To ensure each image's knowledge is sufficiently studied, for all the interleaving data, despite standard filtering, deduping, and other quality control pipeline, we also integrate a data reordering procedure to keep all the image and text in the correct order.  
 
\textbf{OCR Data} 
Optical Character Recognition (OCR) is a widely adopted technique that converts text from images into an editable format. In our model, a robust OCR capability is deemed essential for better aligning the model with human values. Accordingly, our OCR data sources are diverse, ranging from open-source to in-house datasets, encompassing both clean and augmented images, and spanning over single-page and multi-page inputs.

In addition to the publicly available data, we have developed a substantial volume of in-house OCR datasets, covering multilingual text, dense text layouts, web-based content, and handwritten samples. 
Furthermore, following the principles outlined in OCR 2.0~\citep{wei2024general}, our model is also equipped to handle a variety of optical image types, including figures, tables, geometry diagrams, mermaid plots, and natural scene text. We apply extensive data augmentation techniques—such as rotation, distortion, color adjustments, and noise addition—to enhance the model’s robustness. As a result, our model achieves a high level of proficiency in OCR tasks.

In addition to single-page OCR data, we collect and convert a large volume of in-house multi-page OCR data to activate the model's understanding of long documents in the real world. With the help of these data, our model is capable of performing accurate OCR on a single image but can also comprehend an entire academic paper or a scanned book.

\textbf{Knowledge Data} The concept of multimodal knowledge data is analogous to the previously mentioned text pre-training data, except here we focus on assembling a comprehensive repository of human knowledge from diverse sources to further enhance the model's capabilities. 
For example, carefully curated geometry data in our dataset is vital for developing visual reasoning skills, ensuring the model can interpret the abstract diagrams created by humans.

Our knowledge corpus adheres to a standardized taxonomy to balance content across various categories, ensuring diversity in data sources. Similar to text-only corpora, which gather knowledge from textbooks, research papers, and other academic materials, multimodal knowledge data employs both a layout parser and an OCR model to process content from these sources. While we also include filtered data from internet-based and other external resources.

Because a significant portion of our knowledge corpus is sourced from internet-based materials, infographics can cause the model to focus solely on OCR-based information. In such cases, relying exclusively on a basic OCR pipeline may limit training effectiveness. To address this, we have developed an additional pipeline that better captures the purely textual information embedded within images.

\textbf{Agent Data}
For agent tasks, the model's grounding and planning capabilities have been significantly enhanced. In addition to utilizing publicly available data, a platform has been established to efficiently manage and execute virtual machine environments in bulk. Within these virtual environments, heuristic methods were employed to collect screenshots and corresponding action data. This data was then processed into dense grounding formats and continuous trajectory formats. The design of the Action Space was categorized according to Desktop, Mobile, and Web environments. Furthermore, icon data was collected to strengthen the model's understanding of the meanings of icons within software graphical user interfaces (GUIs).
To enhance the model's planning ability for solving multi-step desktop tasks, a set of computer-use trajectories was collected from human annotators, each accompanied by synthesized Chain-of-Thought (Aguvis~\citep{xu2024aguvis}). These multi-step agent demonstrations equip \ourname{} with the capability to complete real-world desktop tasks (on both Ubuntu and Windows).

\textbf{Video Data}  
In addition to image-only and image-text interleaved data, we also incorporate large-scale video data during pre-training, cooldown, and long-context activation stages to enable two directions of essential abilities of our model: first, to understand a long-context sequence dominated by images (\textit{e.g.} hour-long videos) in addition to long text; second, to perceive fine-grained spatio-temporal correspondence in short video clips.

Our video data are sourced from diverse resources, including open-source datasets as well as in-house web-scale video data, and span videos of varying durations.
Similarly, to ensure sufficient generalization ability, our video data cover a wide range of scenes and diverse tasks. We cover tasks such as video description and video grounding, among others.
For long videos, we carefully design a pipeline to produce dense captions. Similar to processing the caption data, we strictly limit the proportion of the synthetic dense video description data to reduce the risk of hallucinations.

\textbf{Text Data}
Our text pretrain corpus directly utilizes the data in Moonlight \cite{liu2025muonscalablellmtraining}, which is designed to provide comprehensive and high-quality data for training large language models (LLMs). It encompasses five domains: English, Chinese, Code, Mathematics \& Reasoning, and Knowledge. 
We employ sophisticated filtering and quality control mechanisms for each domain to ensure the highest quality training data. For all pretrain data, we conducted rigorous individual validation for each data source to assess its specific contribution to the overall training recipe. This systematic evaluation ensures the quality and effectiveness of our diverse data composition.
To optimize the overall composition of our training corpus, the sampling strategy for different document types is empirically determined through extensive experimentation. We conduct isolated evaluations to identify document subsets that contribute most significantly to the model's knowledge acquisition capabilities. These high-value subsets are upsampled in the final training corpus. However, to maintain data diversity and ensure model generalization, we carefully preserve a balanced representation of other document types at appropriate ratios. This data-driven approach helps us optimize the trade-off between focused knowledge acquisition and broad generalization capabilities.
\footnotetext[1]{GPT-4o and GPT-4o-mini results use Omniparser without UIA, according to \cite{bonatti2024windowsagentarenaevaluating}.}
\begin{table}[h!]
\renewcommand{\arraystretch}{1.4}

    \centering
    \footnotesize
    \setlength{\tabcolsep}{1.84pt}
    \begin{tabular}{@{}c l |  c c | c  c c | c c@{}}
    \toprule
    & \multirow{2}{*}{\centering \textbf{Benchmark {\tiny (Metric)}}}  & \multirow{2}{*}{\textbf{GPT-4o}} & \textbf{GPT-} & \textbf{Qwen2.5-} & \textbf{Llama3.2-} & \textbf{Gemma3-}  & \textbf{DeepSeek-}   & \textbf{Kimi-VL-} \\
    & &  & \textbf{4o-mini}  & \textbf{VL-7B} & \textbf{11B-Inst.} & \textbf{12B-IT} & \textbf{VL2}  & \textbf{A3B} \\
    \midrule
    & Architecture  & - & - & Dense & Dense & Dense & MoE & MoE \\
    & \# Act. Params $_\text{(LLM+VT)}$  & - & - & 7.6B+0.7B & 8B+2.6B & 12B+0.4B & 4.1B+0.4B & 2.8B+0.4B \\
    & \# Total Params  & -& - & 8B & 11B & 12B & 28B & 16B \\
    \midrule
    
    \multirow{3}{66pt}{College-level}
    & MMMU$_{\text{val}}$ {\tiny (Pass@1)}  & \gpto{69.1} & \textbf{60.0} & 58.6 & 48 & \underline{59.6} &  51.1 &  {57.0} \\
    & VideoMMMU {\tiny (Pass@1)}  &  \gpto{61.2} &  - & 47.4  & 41.8  & \textbf{57.2}  & 44.4  & \underline{52.6} \\
    & MMVU$_{\text{val}}$ {\tiny (Pass@1)}   & \gpto{67.4}  & \textbf{61.6}  &   50.1 & 44.4  & \underline{57.0} &  52.1 & {52.2} \\
    \midrule
    
    \multirow{5}{66pt}{General}
    & MMBench-EN-v1.1 {\tiny (Acc)}  & \gpto{83.1} & 77.1 & \underline{82.6} & 65.8 & 74.6 & 79.6 &  \textbf{83.1} \\
    & MMStar {\tiny (Acc)}  & \gpto{64.7} & 54.8 & \textbf{63.9} & 49.8 & 56.1 & 55.5 &  \underline{61.3} \\
    & MMVet {\tiny (Pass@1)}   & \gpto{69.1} & \underline{66.9} & \textbf{67.1} & 57.6 & 64.9 & 60.0 &  {66.7}  \\
    & RealWorldQA {\tiny (Acc)}  & \gpto{75.4} & 67.1 & \textbf{68.5} & 63.3 & 59.1 & \underline{68.4} &  {68.1}  \\
    & AI2D {\tiny (Acc)}  & \gpto{84.6} & 77.8 & \underline{83.9} & 77.3 & 78.1 & 81.4 &   \textbf{84.9}  \\
    \midrule
    
    \multirow{1}{66pt}{Multi-image}
    & BLINK {\tiny (Acc)}  & \gpto{68.0} & 53.6 & \underline{56.4} & 39.8 & 50.3 & - & \textbf{57.3} \\
    \midrule
    
    \multirow{2}{66pt}{Math}
    & MathVista {\tiny (Pass@1)}  & \gpto{63.8} & 52.5 & \underline{68.2} & 47.7 & 56.1 & 62.8  & \textbf{68.7} \\
    & MathVision {\tiny (Pass@1)}  & \gpto{30.4} & - & \underline{25.1} & 13.6 & \textbf{32.1} & 17.3  & {21.4} \\
    \midrule
    
    \multirow{2}{66pt}{OCR} 
    & InfoVQA  {\tiny (Acc)}  & \gpto{80.7} & 57.9 & \underline{82.6} & 34.6 & 43.8 & 78.1  & \textbf{83.2}   \\
    & OCRBench {\tiny (Acc)}  & \gpto{815} & 785 & \underline{864} & 753 & 702 & 811  & \textbf{867} \\
    %& DocVQA$_{\text{test}}$  {\tiny (Acc)}  & 91.1 & 72.6 & 95.7 & & 72.7 & 93.3  & 96.0  \\
    \midrule
    
    \multirow{4}{66pt}{OS Agent} 
    & ScreenSpot-V2 {\tiny (Acc)}    & \gpto{18.1}  & - &  \underline{86.8}  & -  & - & - &  \textbf{92.8} \\
    & ScreenSpot-Pro {\tiny (Acc)}  &  \gpto{0.8}  & -  &  \underline{29.0} &  - & - & - &  \textbf{34.5}  \\
    & OSWorld {\tiny (Pass@1)}         &  \gpto{5.03}  & -  &  \underline{2.5} & -  & - & - & \textbf{8.22}    \\   
    & WindowsAgentArena {\tiny (Pass@1)\footnotemark} & \gpto{9.4} & 2.7 & \underline{3.4} & - & - & - & \textbf{10.4}     \\
    \midrule
    
    \multirow{1}{66pt}{Long Document} 
    & MMLongBench-Doc {\tiny (Acc)}    & \gpto{42.8}  & 29.0  &  \underline{29.6} & 13.8  & 21.3 & -  & \textbf{35.1}  \\
    \midrule
    
    \multirow{3}{66pt}{Long Video}
    & Video-MME {\tiny (w/o sub. / w/ sub.)}  & \gpto{71.9/77.2}  & 64.8/68.9  & \underline{65.1/71.6}  &  46.0/49.5  & 58.2/62.1 & - & \textbf{67.8/72.6} \\
    & MLVU$_{\text{MCQ}}$ {\tiny (Acc)}   & \gpto{64.6}  & 48.1  &  \underline{70.2}  &  44.4 & 52.3  & -  & \textbf{74.2} \\
    & LongVideoBench$_{\text{val}}$   & \gpto{66.7}  &  \underline{58.2}  &  56.0 & 45.5  & 51.5  & - & \textbf{64.5}  \\
    %& VideoNIAH & & 64.4 & & & & & & 65.0 \\
    \midrule

    \multirow{3}{66pt}{Video Perception} 
    & EgoSchema$_{\text{full}}$   &  \gpto{72.2} &  - & \underline{65.0}  &  54.3 & 56.9  & 38.5 & \textbf{78.5} \\ 
    & VSI-Bench           &  \gpto{34.0} & -  & \underline{34.2}  & 20.6  & 32.4  & 21.7 & \textbf{37.4} \\ 
    %& TempCompass        & &  73.8 & -  &  71.7 &   &   &  &  \\
    & TOMATO        &  \gpto{37.7} & \underline{28.8}  & 27.6  & 21.5  & 28.6  & 27.2 & \textbf{31.7}  \\

    \bottomrule
    \end{tabular}
    \vspace{1em}
    \caption{Performance of \ourname~against proprietary and open-source efficient VLMs; performance of GPT-4o is also listed in \gpto{gray} for reference. Top and second-best models are in \textbf{boldface} and \underline{underline} respectively. Some results of competing models are unavailable due to limitation of model ability on specific tasks or model context length. }
    \label{tab:short_perf}
    \vspace{-1em}
\end{table}

\subsection{Instruction Data}

\label{sec:3.2}

At this stage, the data is primarily aimed at enhancing the model's conversational abilities and instruction-following capabilities. To cover as many scenarios as possible, we enrich the data across different domains.
For non-reasoning tasks, including chart interpretation, agent grounding, OCR, image-grounded conversations, question-answering, writing, and text processing, we initially construct a seed dataset through human annotation. This seed dataset is used to train a seed model. Subsequently, we collect a diverse set of prompts and employ the seed model to generate multiple responses to each prompt. Annotators then rank these responses and refine the top-ranked response to produce the final version. For reasoning tasks like visual coding, visual reasoning, and math/science problems, where rule-based and model-based verifications are more accurate and efficient than human judgment, we utilize rejection sampling to expand the SFT dataset.
The complete vanilla SFT dataset comprises approximately a 1:1 ratio of text tokens to image tokens.

\begin{figure}[t]
    \centering
    \includegraphics[width=0.8\linewidth]{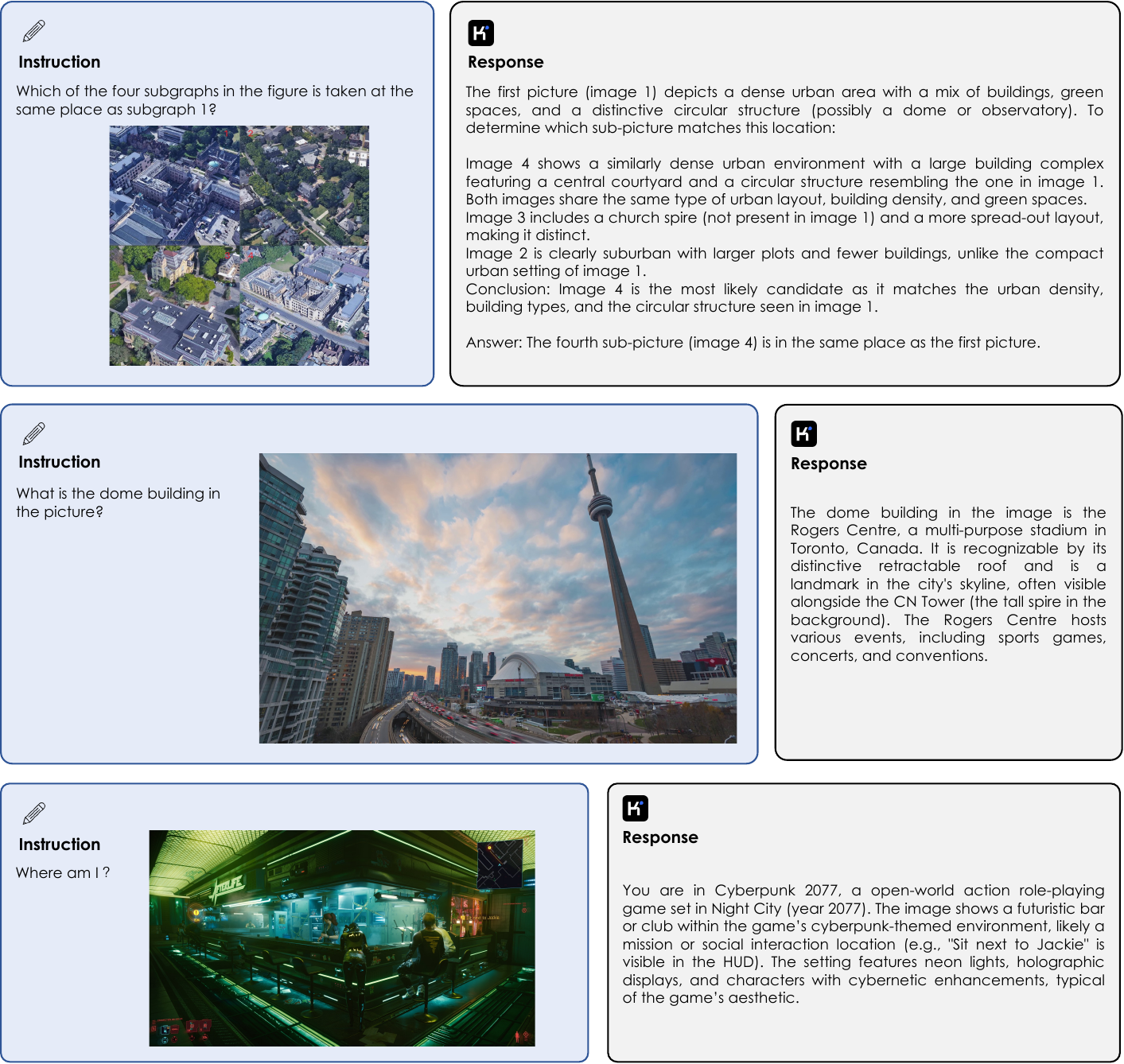}
    \caption{\ourname~exhibits strong visual reasoning capabilities by grounding visual content in spatial, contextual, and cultural knowledge. It accurately identifies matching urban locations based on structural and layout features, interprets scenes from video games like Cyberpunk 2077 using stylistic cues, and recognizes real-world landmarks such as the Rogers Centre in Toronto. }\label{fig:outline}
\end{figure}

\subsection{Reasoning Data}
% \hy{blue}{tianhui here}
Our reasoning data is meticulously constructed for activation and enhancement of the model's multimodal reasoning capabilities during both the long-CoT supervised fine-tuning and reinforcement learning stages.
Through developing a generation pipeline that resembles rejection sampling (RS) and prompt engineering, we collect and synthesize an amount of high-quality long-CoT data.
Specifically, we first assemble a collection of QA data with ground truth annotations that require multi-step reasoning, such as mathematical problem-solving and domain-specific VQA.
Subsequently, we sample multiple detailed reasoning trajectories for each question by leveraging a powerful long-CoT model - Kimi k1.5~\citep{team2025kimi} with curated reasoning prompts.
In rejection sampling, we feed the true labels and model predictions into an off-the-shelf reward model for judgment.
Wrong chain-of-thought responses are filtered out according to the model evaluation as well as some rule-based rewards, thus improving the reasoning data quality.

\section{Evaluation}

We begin by presenting our comprehensive model and conducting a comparative analysis with leading state-of-the-art (SoTA) solutions. Following this introduction, we proceed to assess various sub-capabilities of the model through detailed performance evaluations. This part examines how effectively the model handles different tasks and scenarios, providing insights into its strengths and limitations across diverse functional domains.

\subsection{Comparison to the State-of-the-Art Models}

Table~\ref{tab:short_perf} presents a comprehensive evaluation of \ourname~against state-of-the-art vision-language models across multiple benchmarks. Although having a more parameter-efficient architecture (2.8B+0.4B activated parameters) compared to larger models such as GPT-4o, Llama-3.2-11B-Inst. and Gemma3-12B-IT, \ourname~demonstrates competitive or superior performance in several key areas. Our model employs a Mixture-of-Experts (MoE) architecture similar to DeepSeek-VL2, but outperforms it on most benchmarks with significantly fewer parameters (activated: 2.8B \textit{vs} 4.5B; total: 16B \textit{vs} 28B); it also outperforms Qwen2.5-VL-7B (\textit{actually 8.3B}) on 19 out of 24 benchmarks, though the latter has 2.59$\times$ more activated parameters. The following sections analyze performance across specific domains, which reveals \ourname~'s strengths in OCR, math, agent, long-form content understanding, multi-image and video perception.

\begin{figure}
    \centering
    \includegraphics[width=1.0\linewidth]{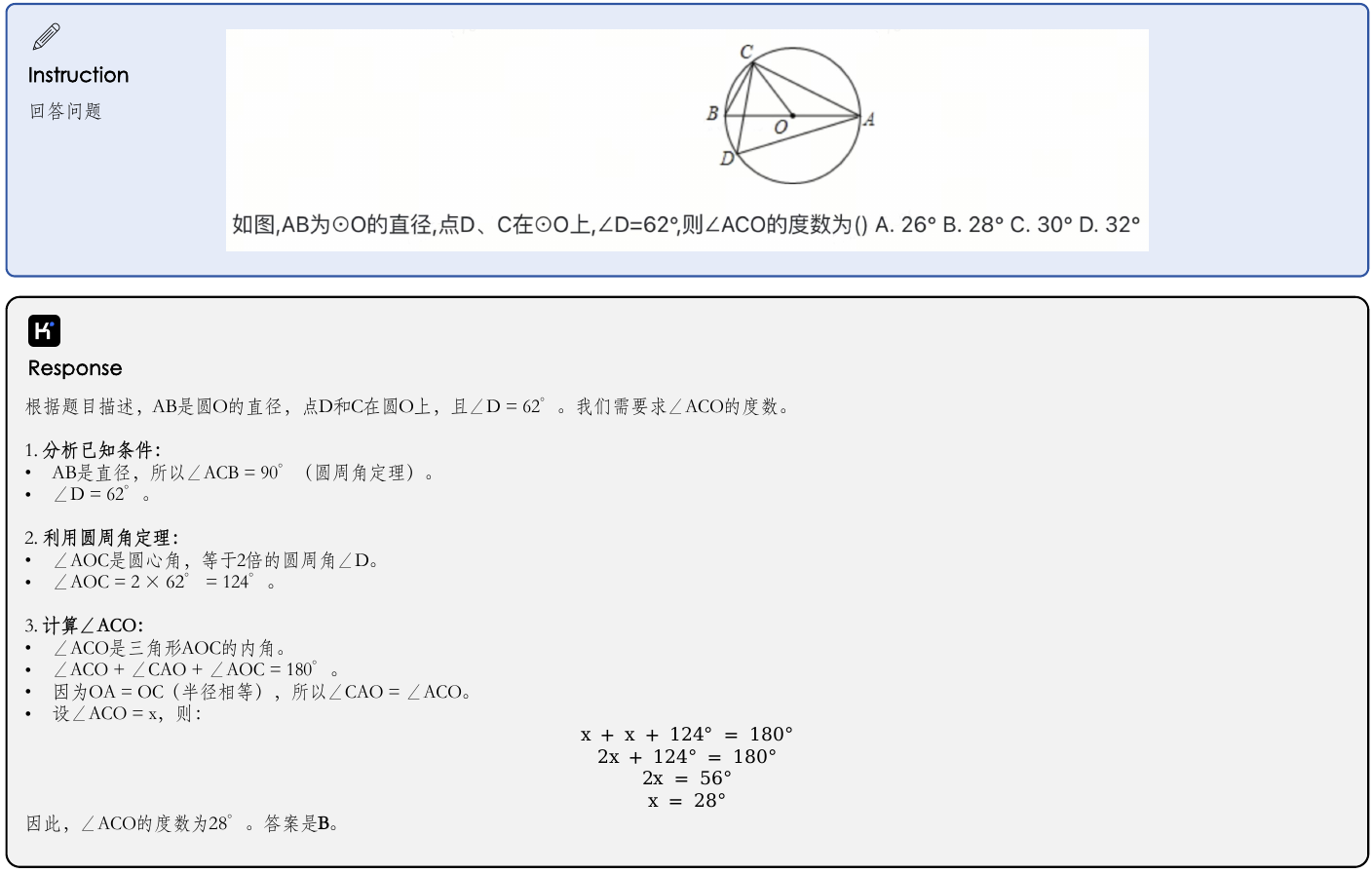}
    \caption{\ourname~demonstrates its capability to perform symbolic reasoning and geometric inference by solving a circle geometry problem step by step. The model analyzes given conditions, applies geometric theorems such as the inscribed angle theorem and properties of triangle angles, and accurately derives the target angle. }
    \label{fig:outline}
\end{figure}

\begin{figure}
    \centering
    \includegraphics[width=1.0\linewidth]{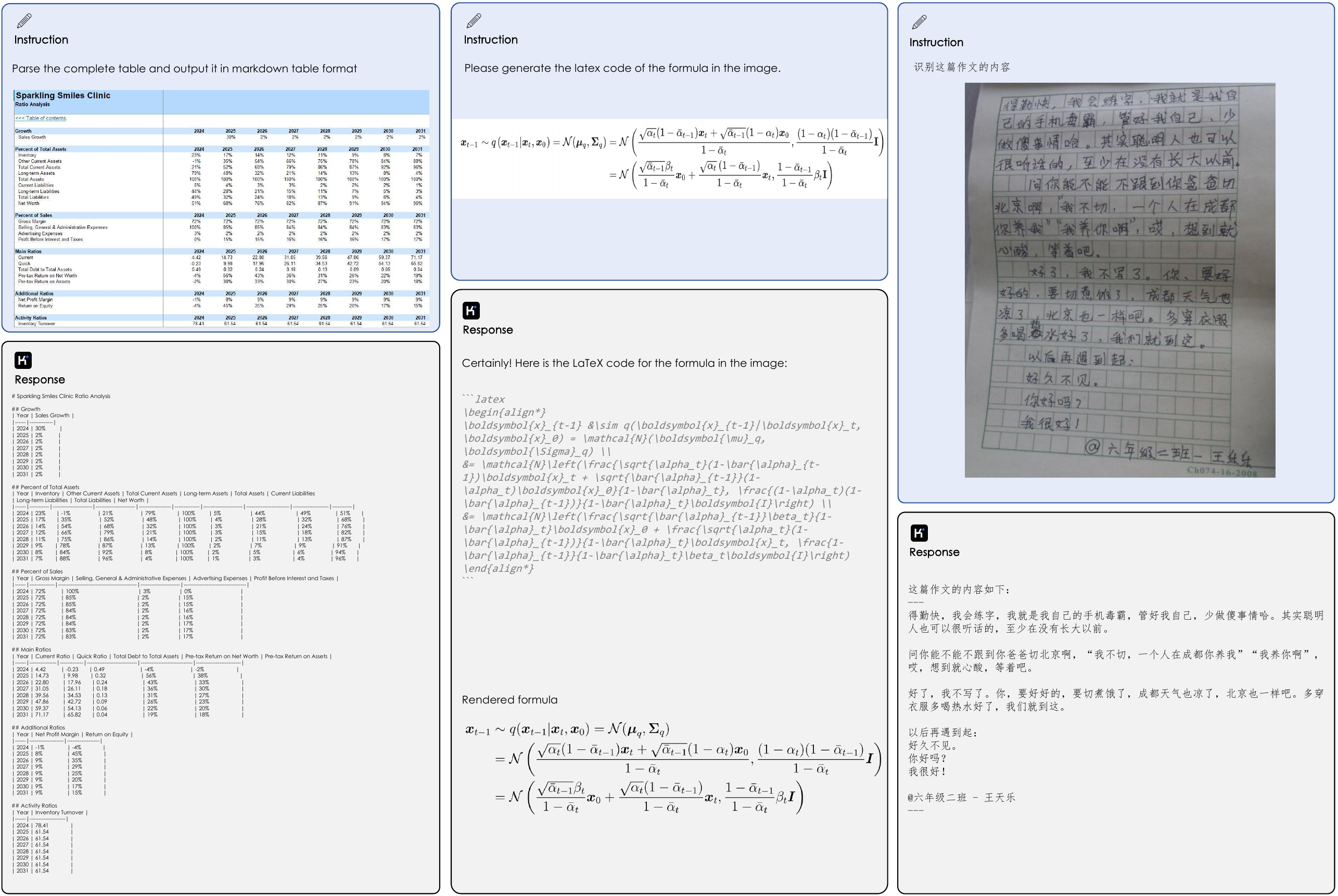}
    \caption{Diverse OCR visualization. \ourname~demonstrates strong OCR capabilities across varied content types, including structured financial tables, complex mathematical formulas, and handwritten Chinese text. The model accurately parses tabular data into markdown, converts formulas to LaTeX, and transcribes handwritten paragraphs with contextual understanding, showcasing its versatility in multimodal text extraction and interpretation.}
    \label{fig:outline}
\end{figure}

\begin{figure}
    \centering
    \includegraphics[width=1.0\linewidth]{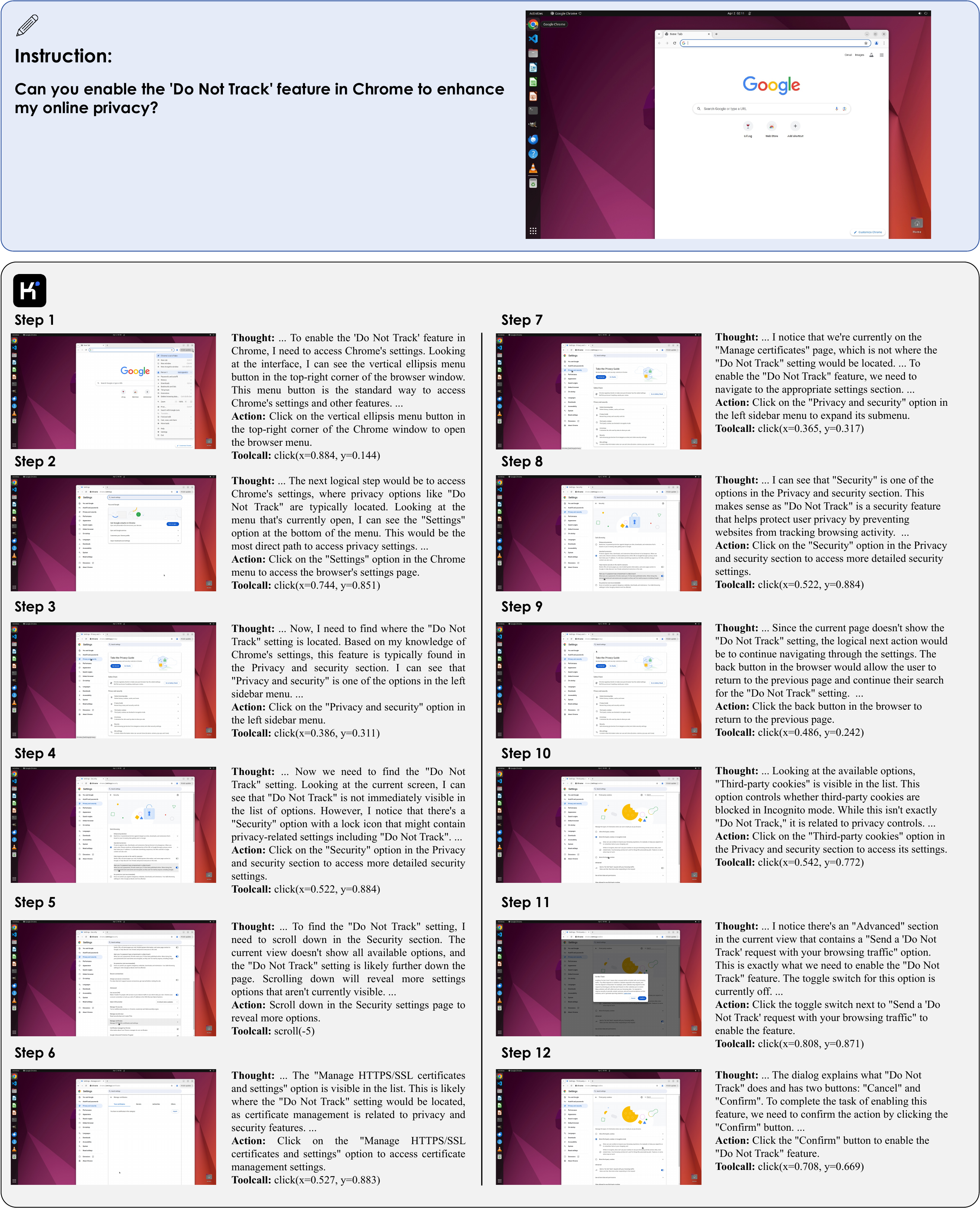}
    \caption{\ourname~is capable of following multi-step reasoning processes to complete complex GUI tasks. In this example, it successfully enables the “Do Not Track” feature in the Chrome browser to enhance online privacy. The agent interprets each screen, identifies relevant UI elements, and performs the appropriate actions sequentially with clear thoughts, actions, and API calls. }
    \label{fig:outline}
\end{figure}

\begin{figure}[h!]
    \centering
    \includegraphics[width=1.0\linewidth]{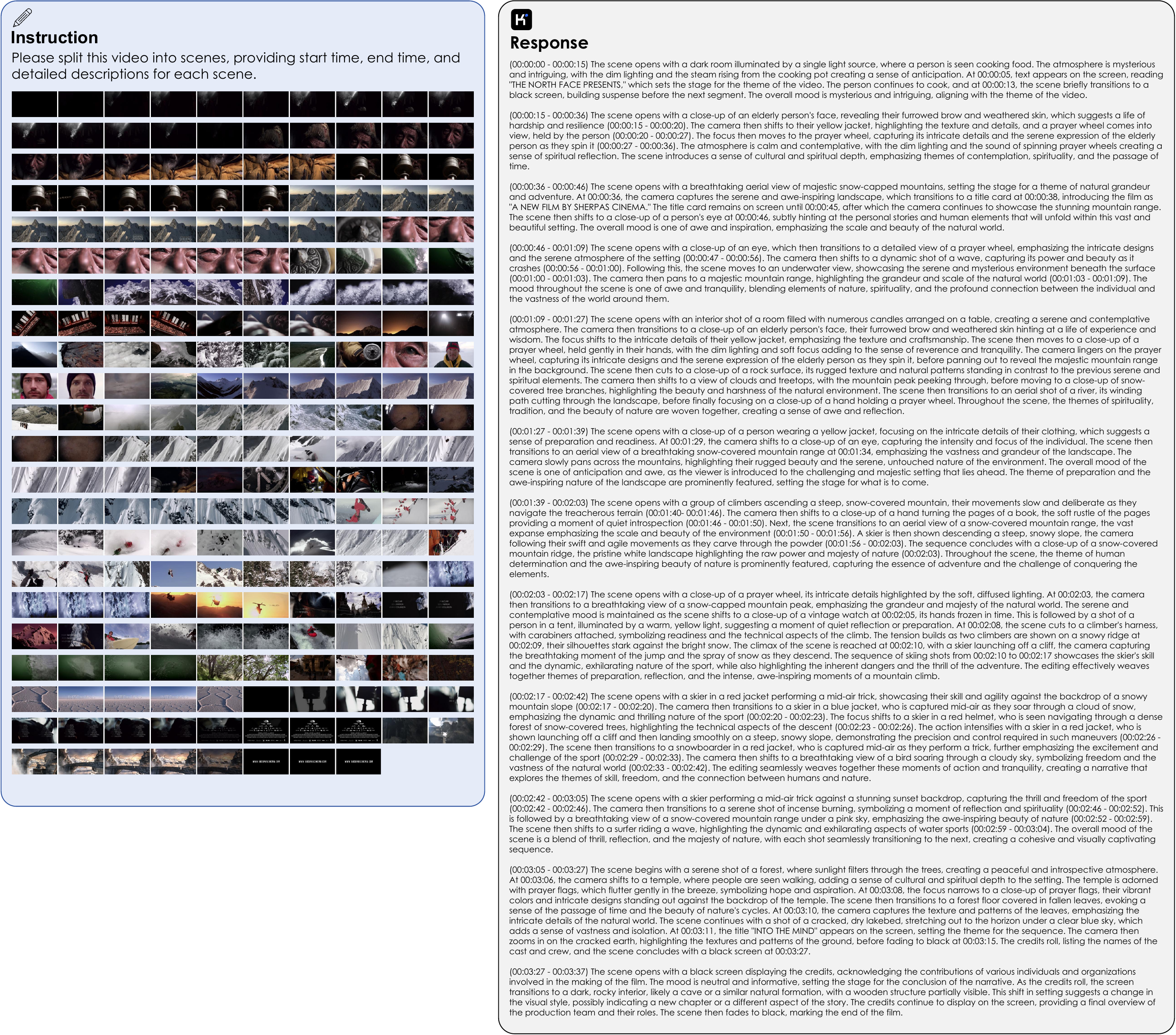}
    \caption{Video scene splitting.
\ourname~processes a long-form video by segmenting it into coherent scenes and providing detailed start/end timestamps along with fine-grained natural language descriptions for each scene.\protect\footnotemark}
    \label{fig:outline}
\end{figure}

\begin{figure}[h!]
    \centering
    \includegraphics[width=1.0\linewidth]{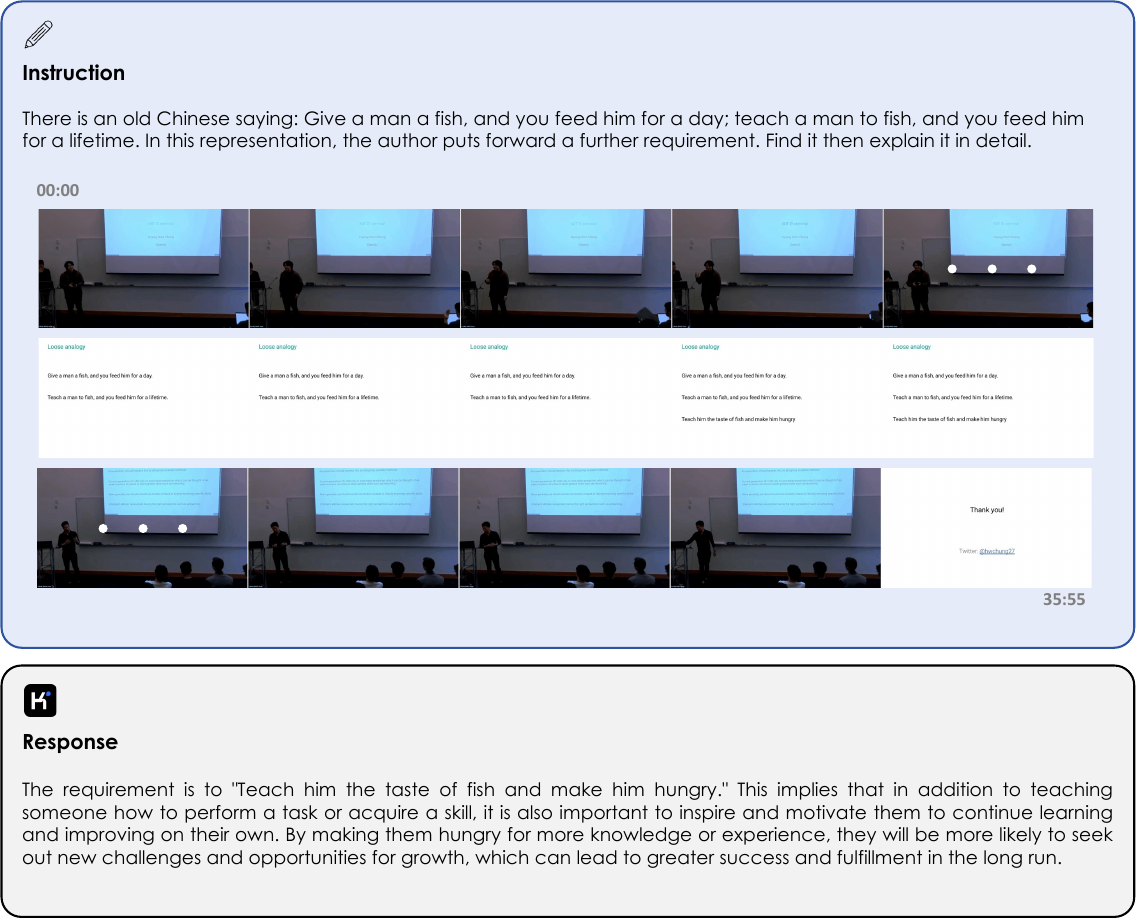}
    \caption{Catching and understanding key details from an hour-long video course.
\ourname~demonstrates its ability to comprehend and interpret instructional video content by analyzing frame sequences and extracting conceptual progression over time. In this case, the model identifies a deepening of the traditional saying ``Teach a man to fish, and you feed him for a lifetime'' into a more nuanced idea: ``Teach him the taste of fish and make him hungry.''\protect\footnotemark}
    \label{fig:outline}
\end{figure}

\subsubsection{College-level Academic Problems}

Our \ourname~model demonstrates competitive performance on college-level academic benchmarks. On MMMU validation set, it achieves a score of 57.0\%, which outperforms DeepSeek-VL2 (51.1\%) and is comparable to Qwen2.5-VL-7B (58.6\%) and even Gemma-3-12B-IT (59.6\%), despite having significantly fewer activated parameters. On video college-level problems, it significantly outperforms Qwen2.5-VL-7B and DeepSeek-VL2, only behind >10B Gemma-3-12B-IT, demonstrating reasonable university-level understanding capabilities compared to larger models. These results indicate that \ourname~effectively balances parameter efficiency with academic reasoning abilities.

\afterpage{\footnotetext[2]{Video source: \url{https://vimeo.com/channels/top/54348266}}}

\subsubsection{General Visual Ability}

\ourname~exhibits strong general visual understanding capabilities across multiple benchmarks. On MMBench-EN-v1.1, it achieves 83.1\% accuracy, outperforming all efficient VLMs in comparison, and performing on par with GPT-4o. For AI2D, our model achieves 84.9\% and surpasses all compared models including GPT-4o (84.6\%). On MMVet, \ourname~scores 66.7\% and ties closely with Qwen2.5-VL-7B (67.1\%) and GPT-4o-mini (66.9\%). For RealWorldQA, it achieves 68.1\%, outperforming Gemma3-12B (59.1\%) and approaching Qwen2.5-VL-7B (68.5\%). These results demonstrate that our model maintains robust general visual understanding despite its compact architecture.

In multi-image reasoning tasks, \ourname~shows promising capabilities with a score of 57.3\% on the BLINK benchmark. This performance surpasses Qwen2.5-VL-7B (56.4\%), GPT-4o-mini (53.6\%), Gemma3-12B-IT (50.3\%), and Llama3.2-11B-Inst. (39.8\%). The ability to reason across multiple images requires understanding spatial and temporal relationships between visual elements, which our model handles effectively with fewer parameters than most competitors. 

\subsubsection{Mathematical Reasoning}

With its relatively small scale, \ourname~also demonstrates strong mathematical reasoning capabilities, particularly on the MathVista benchmark where it achieves 68.7\%, outperforming all compared models including GPT-4o (63.8\%) and Qwen2.5-VL-7B (68.2\%). It indicates our model's exceptional ability to understand and solve mathematical problems presented in visual contexts. On the more challenging MathVision benchmark, due to limited activated parameters, \ourname~outperforms DeepSeek-VL2 and Llama-3.2-11B-Inst., but lags behind Qwen2.5-VL-7B and Gemma-12B-IT. Nevertheless, through RL and test-time scaling, \ourreasoningname~has significantly improved and already on par with 30B-level VLMs (see Table~\ref{tab:long_perf}). These results highlight our model's effectiveness in combining visual perception with mathematical problem-solving, an essential capability for real-world applications.

\afterpage{\footnotetext[3]{Video source: \url{https://www.youtube.com/watch?v=kYWUEV_e2ss}}}

\subsubsection{Document Understanding and OCR}

\ourname~excels in document understanding and OCR tasks across all benchmarks in this category. On InfoVQA, it achieves 83.2\% accuracy, outperforming GPT-4o (80.7\%) and DeepSeek-VL2 (78.1\%). For OCRBench, our model scores 86.7\%, surpassing all other models including GPT-4o-mini (78.5\%) and DeepSeek-VL2 (81.1\%). 
These results demonstrate that our model has exceptional text recognition and document understanding capabilities, making it especially suitable for applications involving document processing and information extraction.

\subsubsection{Agent Grounding and Multi-turn Agent Interaction}

In agent-based tasks, \ourname~demonstrates remarkable performance. On single-step grounding, our model shows strong accuracy, with 92.0\% on ScreenSpot-V2 and 34.5\% on extremely difficult ScreenSpot-Pro (on 4K screens), proving its strong agent grounding abilities. More importantly, it also shows strong multi-step turn agent interaction abilities: For OSWorld, \ourname~reaches 8.22\%, outperforming GPT-4o (5.03\%) and other capable open-source models; On WindowsAgentArena, our model achieves 10.4\%, also surpassing GPT-4o (9.4\%) and others. These results highlight \ourname's exceptional ability to understand and interact with operating system interfaces, suggesting strong potential for applications in automated UI navigation and task execution.

\subsubsection{Long Document and Long Video Understanding}

\ourname~demonstrates competitive performance in long-form content understanding. On MMLongBench-Doc, a challenging benchmark with question-answering on up to 100+ pages, it achieves 35.1\%, outperforming GPT-4o-mini (29.0\%) and Qwen2.5-VL-7B (29.6\%), only behind GPT-4o (42.8\%). For long video understanding, on Video-MME, our model outperforms all efficient VLMs and especially leads on the fairer \textit{w/o subtitle} setting, where models have to find answers from video frames instead of hacking from input subtitles; on \textit{w/ subtitle}setting, it also reaches extraordinary 72.6\% accuracy. On the MCQ subset of MLVU, \ourname~achieves an impressive 74.2\% score, achieving state-of-the-art and surpassing both GPT-4o (64.6\%) and Qwen2.5-VL-7B (70.2\%). For LongVideoBench, it scores 64.5\%, outperforming all compared models except GPT-4o (66.7\%). These results demonstrate \ourname~'s strong capability to understand long-form PDFs and videos.

\subsubsection{Egocentric and Fine-grained Video Perception}

\ourname~also shows strong performance in more nuanced video perception tasks. On EgoSchema full set (\textit{hidden test set}), it achieves 78.5\%, significantly outperforming GPT-4o (72.2\%), Qwen2.5-VL-7B (65.0\%). For VSI-Bench, a very challenging benchmark that requires to understand spatial relationships and correspondences of multiple objects in a video, our model scores 37.4\%, surpassing GPT-4o (34.0\%) and Qwen2.5-VL-7B (34.2\%). In TOMATO that examines fine-grained temporal perception of VLMs, \ourname~reaches 31.7\%, outperforming Qwen2.5-VL-7B (27.6\%) and GPT-4o-Mini (28.8\%). These results demonstrate our model's strong capability to understand dynamic visual content, track objects over time, and interpret complex actions in video sequences, making it well-suited for applications requiring temporal visual understanding.

\subsection{\texttt{Kimi-VL-A3B-Thinking}: A Reasoning Extension of \ourname}

Furthermore, we conduct a reasoning extension to empower \ourname~to reason with CoT and present a long-thinking version of the model, \textbf{\ourname-Thinking}, through long-CoT activation and reinforcement learning.
We validate its superior performance on several image benchmarks, as shown in Table~\ref{tab:long_perf}.

\begin{table}[t!]
\renewcommand{\arraystretch}{1.33}
\setlength{\tabcolsep}{2.5pt}
    \centering
    \footnotesize
    \begin{tabular}{l|cccccc|ccc|cc}
    \toprule
    & \multicolumn{6}{c|}{\textbf{Non-Thinking Model}} & \multicolumn{5}{c}{\textbf{Thinking Model}} \\
    \textbf{Benchmark {\tiny (Metric)}}  & \multirow{2}{*}{\textbf{GPT-4o}} & \textbf{GPT-}   &\multicolumn{2}{c}{\textbf{Qwen2.5-VL-}}&\multicolumn{2}{c|}{\textbf{Gemma-3-}}  & \textbf{o1-} & \textbf{QVQ-72B-} & \textbf{Kimi-} & \multicolumn{2}{c}{\textbf{\ourname-A3B-}} \\
    & & \textbf{4o-mini} & \textbf{72B} & \textbf{7B} & \textbf{27B} & \textbf{12B} & \textbf{1217} & \textbf{Preview} & \textbf{k1.5} & \textbf{Thinking} & \textbf{Thinking-2506} \\
    \midrule
    MathVision (full) {\tiny (Pass@1)} & 30.4 & - & 38.1 & 25.1 & 35.5 & 32.1 & - & 35.9 & 38.6 & 36.8 & \textbf{56.9} \\
    MathVista (mini) {\tiny (Pass@1)} & 63.8 & 56.7 & 74.8 & 68.2 & 62.3 & 56.4 & 71.0 & 71.4 & 74.9 & 71.3 & \textbf{80.1} \\
    MMMU (val) {\tiny (Pass@1)} & 69.1 & 60.0 & 74.8 & 58.6 & 64.8 & 59.6  & \textbf{77.3} & 70.3 & 70.0 & 61.7 & 64.0 \\
    MMMU-Pro (avg) {\tiny (Pass@1)} & \textbf{51.7} & 37.6 & 51.1 & 38.1 & - & 32.1 & - & - & - & 43.0 & 46.3  \\
    VideoMMMU {\tiny (Pass@1)} & 61.1 & - & 60.2 & 47.0 & 61.8 & 57.2  & - & -  & - & 55.5 & \textbf{65.2} \\
    \bottomrule
    \end{tabular}
    \vspace{1em}
    \caption{Performance of \ourreasoningname{} and \ourreasoningname-2506 on multimodal reasoning benchmarks. The metrics evaluated include MathVista (mini), MMMU (val), MMMU-Pro (average), MathVision (full) and VideoMMMU, with results expressed in Pass@1. The \ourreasoningname{}-2506 performs well in most cases, showcasing the enhanced reasoning and processing capabilities of the \textit{``thinking''} variant across different domains and scales.}
    \label{tab:long_perf}
\end{table}

\begin{figure}[t!]
    \centering
    \includegraphics[width=1.0\linewidth]{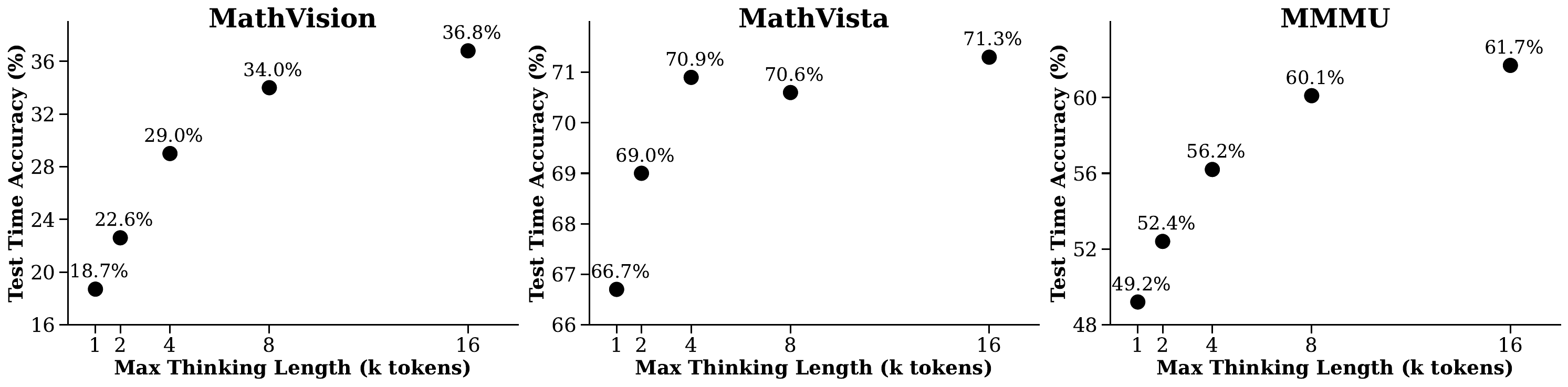}
    \caption{Test-time accuracy when scaling the max thinking token length of our \textbf{Kimi-VL-Thinking} model.}
    \label{fig:test_time_accuracy}
    \vspace{-1em}
\end{figure}

\ourreasoningname~significantly improves over the base \ourname~model, with gains of 2.6\% on MathVista, 4.7\% on MMMU, and 15.4\% on MathVision, demonstrating its capability to leverage test-time computation for deeper reasoning and better handling of complex multimodal queries. In Table~\ref{tab:long_perf}, \ourname-Thinking further outperforms or rivals state-of-the-art thinking and non-thinking models: achieving 71.3\% on MathVista, outperforming GPT-4o (63.8\%) and GPT-4o-mini (56.7\%); scoring 61.7\% on MMMU, surpassing GPT-4o-mini (60.0\%) and Qwen2.5-VL-7B (58.6\%); and reaching 36.8\% on MathVision, exceeding GPT-4o (30.4\%) and Gemma-3-27B-IT (35.5\%), even QVQ-72B (35.9\%). While marginally behind some larger-scale models on select benchmarks, \ourname-Thinking accomplishes these results with only 3B activated parameters—orders of magnitude fewer than its counterparts—underscoring its strong efficiency and effectiveness in multimodal reasoning.

Our Kimi-VL-Thinking model also exhibits strong test-time scaling properties, as shown in Figure~\ref{fig:test_time_accuracy}. Specifically, increasing the max thinking token length at inference time consistently improves test-time accuracy across all three benchmarks. For example, on \textbf{MathVision}, the accuracy rises steadily from 18.7\% at 1k tokens to 36.8\% at 16k tokens, and similar upward trend is also observed on \textbf{MMMU}, indicating that the model is able to utilize longer reasoning chains for better performance.
However, not all benchmarks benefit equally from longer thinking lengths. On \textbf{MathVista}, performance saturates early, with accuracy reaching 70.9\% at 4k tokens and no further significant gains observed as the token length increases to 16k. It suggests that for this task, the necessary reasoning depth is already captured within a relatively short context, and additional computation does not yield further improvements. 

\subsection{\texttt{Kimi-VL-A3B-Thinking-2506}: From Reasoning Extension to Integrated Thinking Model}

\begin{table*}[htbp]
  \centering
  \renewcommand{\arraystretch}{1.1}
  \caption{Performance of Kimi-VL-A3B-Thinking-2506 on multimodal benchmarks that do not require extensive reasoning.}
  \label{tab:2506_non_reasoning_bmks}

  % This command creates a stacked header, allowing line breaks.
  \newcommand{\stackheader}[1]{\textbf{\begin{tabular}{@{}c@{}}#1\end{tabular}}}

  % \resizebox is used to ensure the wide table fits within the text width of the page.
  \resizebox{\textwidth}{!}{%
  % Column definitions:
  % @{} removes extra padding at the edges.
  % l for the first column (left-aligned).
  % >{\columncolor{gray}}c applies the gray background to the second column (center-aligned).
  % ccccc for the remaining five columns (center-aligned).
  \begin{tabular}{@{}lc ccccc@{}}
    \toprule
    \textbf{Benchmark (Metric)} & 
    \textbf{GPT-4o} & 
    \stackheader{Qwen2.5-\\VL-7B} & 
    \stackheader{Gemma3-\\12B-IT} & 
    \stackheader{Kimi-VL-A3B-\\Instruct} & 
    \stackheader{Kimi-VL-A3B-\\Thinking} & 
    \stackheader{Kimi-VL-A3B-\\Thinking-2506} \\
    \midrule
    \multicolumn{7}{@{}l}{\textit{General Multimodal}} \\
    MMBench-EN-v1.1 {\tiny (Acc)} & \gpto{83.1} & 83.2 & 74.6 & 82.9 & 76.0 & \textbf{84.4} \\
    RealWorldQA {\tiny (Acc)}     & \gpto{75.4} & 68.5 & 59.1 & 68.1 & 64.0 & \textbf{70.0} \\
    OCRBench {\tiny (Acc)}        & \gpto{815}  & 864  & 702  & 864  & 864  & \textbf{869}  \\
    MMStar {\tiny (Acc)}          & \gpto{64.0} & 63.0 & 56.1 & 61.7 & 64.2 & \textbf{70.4} \\
    MMVet {\tiny (Acc)}           & \gpto{69.1} & 67.1 & 64.9 & 66.7 & 69.5 & \textbf{78.1} \\
    \midrule
    \multicolumn{7}{@{}l}{\textit{Video}} \\
    MMVU$_\text{val}$ {\tiny (Pass@1)}         & \gpto{67.4} & 50.1 & 57.0 & 52.7 & 53.0 & \textbf{57.5} \\
    Video-MME (w/ sub.) {\tiny (Acc)}   & \gpto{77.2} & 71.6 & 62.1 & \textbf{72.7} & 66.0 & 71.9 \\
    \midrule
    \multicolumn{7}{@{}l}{\textit{OS-Agent Grounding}} \\
    ScreenSpot-Pro {\tiny (Acc)}  & \gpto{0.8}  & 29.0 & ---  & 35.4 & ---  & \textbf{52.8} \\
    ScreenSpot-V2 {\tiny (Acc)}   & \gpto{18.1} & 84.2 & ---  & \textbf{92.8} & ---  & 91.4 \\
    OSWorld-G {\tiny (Acc)}       & -    & 31.5 & ---  & {41.6} & ---  & \textbf{52.5} \\
    \midrule
    \multicolumn{7}{@{}l}{\textit{Long Document}} \\
    MMLongBench-Doc {\tiny (Acc)} & \gpto{42.8} & 29.6 & 21.3 & 35.1 & 32.5 & \textbf{42.1} \\
    \bottomrule
  \end{tabular}%
  }
\end{table*}

While Kimi-VL-A3B-Thinking shows excellent thinking abilities on hard reasoning tasks, we further provide the updated \textbf{{Kimi-VL-A3B-Thinking-2506}}\footnote{Tech Blog: \url{https://huggingface.co/blog/moonshotai/kimi-vl-a3b-thinking-2506}}, a new reasoning variant that is not only smarter, but integrates key abilities of Kimi-VL-A3B-Instruct (perception, video, long-document, and OS-agent abilities) into this thinking model.

\ourreasoningname-2506 significantly improves reasoning efficiency while reducing token consumption. As shown in Table \ref{tab:long_perf}, \ourname-Thinking-2506 achieves 56.9\% on MathVision (+20.1\% improvement on original \ourreasoningname), 80.1\% on MathVista (+8.4\%), 46.3\% on MMMU-Pro (+3.2\%), and 64.0\% on MMMU (+2.1\%), demonstrating non-trivial gains across multiple reasoning benchmarks. Meanwhile, while solving these hard reasoning problems, the 2506 version reduces the average output token length by around 20\% (\textit{e.g.}, 2.9K~$\to$~2.4K on MMMU-val and 5.8K~$\to$~4.4K on MathVision), facilitating it to be more efficient and user-friendly for practical deployments.

Beyond extensive reasoning tasks, \ourreasoningname~demonstrates stronger visual perception capabilities (Table \ref{tab:2506_non_reasoning_bmks}). Compared to the previous non-thinking variant (Kimi-VL-A3B-Instruct), Kimi-VL-A3B-Thinking-2506 achieves competitive or superior results on general multimodal understanding benchmarks: 84.4\% on MMBench-EN-v1.1, 70.4\% on MMStar, 70.0\% on RealWorldQA, and 78.4\% on MMVet, underscoring its broader competence in vision-language tasks. In terms of token efficiency, the 2506 version only requires in average 180 tokens per answer when solving MMBench, 1/3 compared to the previous thinking model while improving 8.4\% accuracy.

Kimi-VL-A3B-Thinking-2506 also extends its reasoning ability to video and long-context domains. It establishes new state-of-the-art results among open-source models on VideoMMMU (65.2\%, 4\% better than GPT-4o), a challenging video reasoning benchmark; it also maintains robust general video understanding performance with 71.9\% on Video-MME, matching the long video understanding ability of Kimi-VL-A3B-Instruct. It also scores 42.1\% (first open-source model matching GPT-4o) on MMLongBench-Doc (Table \ref{tab:2506_non_reasoning_bmks}), a 10\% improvement over the previous thinking model and 7\% over the previous instruct model, demonstrating its robust ability on broader long-form visual inputs. 

As mentioned in the method part, the continual training on MoonViT (3.2 million max input pixels) of Kimi-VL-A3B-Thinking-2506 leads to substantial improvements on \textbf{high-resolution} perception and OS grounding benchmarks, achieving 83.2\% on V* Benchmark (without external tools), 52.8\% on ScreenSpot-Pro, and 52.5\% on OSWorld-G (full set with refusal samples), showing huge improvements over both previous variants. We hope that this high-resolution multimodal reasoning model brings about interesting new capabilities in the real world.

\section{Conclusion, Limitation, and Future Work}

We introduce \ourname, a VLM designed with a balanced approach to cover both multimodal and text-only pre-training/post-training, underpinned by an MoE-based architecture for scalable efficiency. Its 128K extended context window enables precise retrieval in lengthy texts and videos, while the native-resolution encoder MoonViT helps maintain high accuracy with low computational overhead in ultra-high-resolution visual tasks. Additionally, \ourname-Thinking facilitates effective long-chain reasoning in complex image and video inference. Overall, \ourname~demonstrates robust adaptability and efficiency across multimodal, long-context, and high-resolution tasks, indicating substantial potential for future research and industrial applications.

\noindent
However, \ourname~still faces several challenges:
\begin{enumerate}
    \item Although the current model size performs effectively for many standard tasks, it remains too limited to address highly specialized or domain-specific problems, or problems that are strongly dependent on language abilities, restricting \ourname's ability to handle extremely complex scenarios.
    \item While the reasoning capability is already strong for typical use cases, it has yet to reach its theoretical upper bound, particularly for intricate tasks requiring multi-step inference or deeper contextual understanding.
    \item Despite providing a 128K extended context window, due to limited parameters in its attention layers (which is only comparable to a 3B model), its long-context abilities is still insufficient for certain advanced applications that involve extremely long sequences or high-volume contextual information.
\end{enumerate}

\noindent

In the future, we will tackle these challenges by scaling up the model size, expanding pre-training data, and enhancing post-training algorithms. Our next steps include optimizing \ourname~and releasing larger versions, as well as refining post-training and test-time scaling mechanisms for a better thinking model. These efforts will pave the way for more advanced applications in both research and industry.

% \bibliographystyle{unsrtnat}
% \bibliography{references}  %%% Uncomment this line and comment out the ``thebibliography'' section below to use the external .bib file (using bibtex) .
\printbibliography[title={References}]

\newpage
\appendix

\section*{Appendix}

\section{Contributions}

\begin{multicols}{2}
\noindent
\textbf{Core Contributors} \\

Bohong Yin \\
Bowei Xing \\
Cheng Chen \\
Chu Wei \\
Dehao Zhang \\
Dongliang Wang \\
Haoning Wu$^*$ \\
Haotian Yao \\
Haoyu Lu$^*$ \\
Hao Yang \\
Kun Ouyang\\
Lin Sui \\
Xinyuan Wang${^\#}$ \\
Xinyu Zhou \\
Yang Li \\
Y. Charles$^*$ \\
Yiping Bao \\
Yimin Chen \\
Yuanxin Liu\\
Yuxin Wu \\
Zaida Zhou \\
Zhaowei Li \\
Zhiqi Huang \\
Zhilin Yang \\
Ziwei Chen \\

\textbf{Contributors} \\

Angang Du \\
Bowen Qu \\
Bowen Wang${^\#}$ \\
Chenlin Zhang \\
Chenzhuang Du \\
Congcong Wang \\
Dikang Du \\
Enming Yuan \\
Enzhe Lu \\
Fang Li \\
Flood Sung \\
Guangda Wei \\
Guokun Lai \\
Han Zhu \\
Hao Ding \\
Hao Hu \\
Hao Zhang \\
Heng Wang \\
Hongcheng Gao \\
Huabin Zheng \\
Jiaming Li \\
Jianlin Su \\
Jianzhou Wang \\
Jiaqi Deng${^\#}$ \\
Jiezhong Qiu \\
Jin Xie \\
Jinhong Wang \\
Jingyuan Liu \\
Junjie Yan \\
Liang Chen \\
Longhui Yu \\
Mengfan Dong \\
Mengnan Dong \\
Nuo Xu \\
Pengyu Cheng \\
Qizheng Gu \\
Runjie Zhou \\
Shaowei Liu \\
Sihan Cao \\
Tao Yu$^\#$ \\
Tianhui Song \\
Tongtong Bai \\
Weiran He \\
Wei Song \\
Weixiao Huang \\
Weixin Xu \\
Xiaokun Yuan \\
Xingzhe Wu \\
Xingcheng Yao \\
Xinhao Li \\
Xinxing Zu \\
Yangyang Hu \\
Yan Zhong \\
Yanru Chen \\
Yibo Miao \\
Yejie Wang \\
Yibo Liu \\
Yidao Qin \\
Yiqin Wang \\
Yongsheng Kang \\
Yuhao Dong \\
Yulun Du \\
Yuzhi Wang \\
Yuzi Yan \\
Zhejun Jiang \\
Zheng Zhang \\
Zihao Huang \\
Zijia Zhao \\
Zongyu Lin \\

\end{multicols}

* Project lead(s).\\
\# The University of Hong Kong, Moonshot.ai \\
The listing of authors is in alphabetical order based on their first names.

\section{Evaluation Details}
\label{sec:appendix_eval_detail}

\subsection{Image Benchmark}

\textbf{MMMU}~\citep{yue2024mmmu} encompasses a carefully curated collection of 11.5K multimodal questions sourced from college exams, quizzes, and textbooks. These questions span six major academic fields: Art \& Design, Business, Science, Health \& Medicine, Humanities \& Social Science, and Tech \& Engineering.

\textbf{MMBench-EN-v1.1}~\citep{MMBench} is a fine-grained benchmark that contains 2974 multiple-choice questions, covering 20 ability dimensions. It incorporate perception and reasoning as the top-level ability dimensions in its ability taxonomy, leading to different levels of evaluation in various ability dimensions.

\textbf{MMStar}~\citep{chen2024mmstar} is an elite vision-indispensable multimodal benchmark comprising 1,500 challenge samples meticulously selected by humans. It is designed to benchmark 6 core capabilities and 18 detailed axes, aiming to evaluate the multimodal capacities of LVLMs with a carefully balanced and purified selection of samples.

\textbf{MMVet}~\citep{yu2024mmvet} is designed based on the insight that the intriguing ability to solve complicated tasks is often achieved by a generalist model being able to integrate different core vision-language  capabilities. It defines 6 core VL capabilities and examines the 16 integrations of interest derived from the capability combination.

\textbf{RealWorldQA}~\citep{realworldQA} is a benchmark designed to evaluate the real-world spatial understanding capabilities of multimodal  models. It assesses how well the models comprehend physical environments. The benchmark consists of over 700 images, each accompanied by a question and a verifiable answer, and these images are drawn from various real-world scenarios.

\textbf{AI2D}~\citep{kembhavi2016ai2d} is a dataset of over 5000 grade school science diagrams with over 150000 rich annotations, their ground truth syntactic parses, and more than 15000 corresponding multiple choice questions.

\textbf{MathVision}~\citep{wang2024measuring} is a carefully curated collection of 3,040 high-quality mathematical problems with visual contexts that are sourced from real math competitions. It covers 16 distinct mathematical disciplines and is graded across 5 levels of difficulty. This dataset offers a comprehensive and diverse set of challenges, making it ideal for evaluating the mathematical reasoning abilities of LMMs.

\textbf{MathVista}~\citep{lu2023mathvista} is a benchmark that integrates challenges from a variety of mathematical and visual tasks, demanding participants to exhibit fine-grained, deep visual understanding along with compositional reasoning to successfully complete the tasks.

\textbf{BLINK}~\citep{fu2024blink} is a benchmark designed to evaluate multi-image visual cognition, encompassing tasks related to depth relationships, feature matching, digital forensics, and spatiotemporal reasoning. It features a diverse set of multi-image perceptual similarity tasks, validated through standardized protocols.

\textbf{InfoVQA}~\citep{mathew2022infographicvqa} is a dataset specifically designed to assess models' capabilities in interpreting and reasoning with complex infographics that integrate text, graphics, and visual elements. Model performance on this dataset is evaluated using the ANLS metric on the test set.

\textbf{OCRBench}~\citep{liu2023hidden} evaluates the OCR capabilities of MLLMs across five tasks: text recognition, scene text VQA, document VQA, key information extraction, and handwritten math expression recognition. The benchmark is scored out of a maximum of 1000 points.

\subsection{Video and Long Document Benchmark}
\textbf{VideoMMMU}~\citep{arxiv2025videommmu} is a video benchmark designed to evaluate the college-level knowledge acquisition capabilities of large multimodal models. It consists of 300 expert-level videos and 900 human-annotated questions. The videos span six diverse academic disciplines: Art, Humanities, Medicine, Business, Science, and Engineering. The questions are structured to align with three cognitive stages: Perception, Comprehension, and Adaptation.

\textbf{MMVU}~\citep{arxiv2025mmvu} is a video benchmark designed to evaluate the expert-level video understanding ability. The benchmark contains 3,000 expert-annotated questions over 1,529 videos, which span 27 subjects from four core disciplines: Science, Healthcare, Humanities \& Social Sciences, and Engineering.

\textbf{Video-MME}~\citep{arxiv2024videomme} is a video benchmark that consists of 900 manually selected videos~(totaling 254 hours length), and 2,700 QA pairs. The videos, varying in duration, are categorized into 30 fine-grained classes across six diverse domains: Knowledge, Film \& Television, Sports Competition, Artistic Performance, Life Record, and Multilingual content. Evaluations are conducted under two different settings: with and without subtitles.

\textbf{MLVU}~\citep{arxiv2024mlvu} is designed to evaluate the model performance in comprehending long videos from multiple aspects. It consists of 1,730 videos along with 3,102 corresponding question-answer pairs~(2,593 in dev set and 509 in test set). Videos of this benchmark are collected from multiple scenarios, including Sport, Ego-centric, Life Record, Tutorial, etc. The close-ended task set of MLVU comprises 7 different tasks: Action Order, Action Count, Topic Reasoning, Anomaly Recognition, Plot QA, Ego Reasoning, and Needle QA.

\textbf{LongVideoBench}~\citep{nips2024longvideobench} is a video question-answering benchmark designed to evaluate the long-form multimodal perception and relation capability of large multimodal models. The benchmark includes 3,763 web-collected videos spanning various lengths and themes, along with their corresponding subtitles. It includes 6,678 human-annotated multiple-choice questions, distributed across 17 fine-grained categories, which accesses different aspects of video-language understanding.

\textbf{EgoSchema}~\citep{nips2023egoschema} is a video benchmark designed to evaluate the long-form video understanding capabilities within the ego-centric scenario. Derived from Ego4D~\citep{cvpr2022ego4d}, the benchmark comprises over 5,031 multiple choice question-answer pairs spanning more than 250 hours real-world videos with a semi-automatic data pipeline. 

\textbf{VSI-Bench}~\citep{arxiv2024vsibench} is designed to evaluate the visual-spatial comprehensive capabilities of large multimodal models. It consists of over 5,000 question-answer pairs across around 290 real indoor-scene videos. 

\textbf{TOMATO}~\citep{iclr2025tomato} is a video benchmark comprises 1,484 human-annotated question-answer pairs and 1,417 videos. TOMATO focuses on evaluating the temporal reasoning capabilities of large multimodal models, including action counting, direction prediction, rotation analysis, shape \& trend detection, velocity \& frequency estimation, and visual cue interpretation.

\subsection{Agent Benchmark}
\textbf{ScreenSpot V2}~\citep{wu2024osatlas} is an enhanced version of the ScreenSpot \citep{cheng2024seeclick} benchmark, which focuses on evaluating the performance of GUI grounding models across multiple platforms, including web, desktop, and mobile interfaces. This updated version addresses several issues identified in the original ScreenSpot dataset, such as incorrect or ambiguous annotations, spelling mistakes, and mislabeled bounding boxes.

\textbf{ScreenSpot Pro}~\citep{li2025screenspotpro} is a benchmark for evaluating GUI grounding in high-resolution, complex UI environments. It contains 1,581 real-world, high-resolution images and expert-annotated tasks from diverse professional domains. Including domain-specific interface conventions that challenge models to understand professional-grade interfaces beyond consumer applications.

\textbf{OSWorld}~\citep{xie2024osworld} is a pioneering scalable, real computer environment designed for multimodal agents, facilitating task setup, execution-based evaluation, and interactive learning across multiple operating systems, including Ubuntu, Windows, and macOS. It serves as a unified platform for evaluating open-ended computer tasks that involve arbitrary applications, addressing the limitations of existing benchmarks that often lack interactive environments or are confined to specific applications or domains.

\textbf{WindowsAgentArena}~\citep{bonatti2024windowsagentarenaevaluating} is a benchmark designed to evaluate multimodal agents in realistic Windows environments. Built on the OSWorld framework, it allows agents to interact with a full range of applications and web tools. The benchmark is scalable and can complete evaluations in under 20 minutes on Azure. It offers insights into agent performance, highlighting the potential for future research in agent development and task automation.

\end{document}